\documentclass{article}

\usepackage[main, final]{neurips_2026}

\usepackage[utf8]{inputenc} 
\usepackage[T1]{fontenc}    
\usepackage{hyperref}       
\usepackage{url}            
\usepackage{booktabs}       
\usepackage{amsfonts}       
\usepackage{nicefrac}       
\usepackage{microtype}      
\usepackage{xcolor}         
\usepackage{kky}

\usepackage{amsmath}
\usepackage{todonotes}
\usepackage[inoutnumbered, algoruled, vlined]{algorithm2e}
\usepackage{cleveref}

\title{Improved regret bounds in tabular RL under local 
differential privacy}

\author{%
  Hugo Richard \\
  Joint team Fairplay, Criteo AI Labs, ENSAE CREST, INRIA\\
  France\\
  \texttt{research.hugo.richard@gmail.com} \\
}

\begin{document}

\maketitle

\begin{abstract}
We study regret minimization for time-homogeneous
episodic tabular reinforcement learning with $S$ states and $A$ actions under   
non-interactive $\epsilon$-local differential privacy (LDP). 
In each episode $k \in [K]$, the decision-maker sends a policy to a user, the
user executes the policy for $H$ steps, and sends an $\epsilon$-locally
differentially private message back to the decision-maker. When $H=1$, the problem reduces to an LDP contextual bandit 
with $S$ contexts, $A$ arms and horizon $K$. 
We prove that the minimax regret is $\min\bigg(K,  \frac{ \sqrt{\max( S, \exp(\epsilon) + 1) S A K (\exp(\epsilon) +
1)}}{\exp(\epsilon) - 1}\bigg)$, up to constants and a $\sqrt{\log(A)}$
factor.
The upper bound is achieved by an adaptation of EXP3 and a unary encoding mechanism. 
When $H > 1$, and for small $\epsilon$, we show a lower bound in
$\Omega(HS\frac{\sqrt{A K}}{\epsilon})$ and an 
upper bound in $\tilde{O}(H^2 S^2 \frac{\sqrt{ AK}}{\epsilon})$.
The upper bound is achieved by an algorithm based on UCBVI and a unary
encoding mechanism. For small $\epsilon$, these bounds improve the previously known lower bound by a factor
$\sqrt{S}$ and the previously known upper bound by a factor of $H \sqrt{A}$. 
\end{abstract}  
 
\section{Introduction}
\emph{Reinforcement learning} (RL) is the problem of learning to make
decisions through interaction.  Typical applications of
RL include recommender systems,
online advertising, healthcare, routing or
robotics~\cite{koberReinforcementLearningRobotics2013,afsarReinforcementLearningBased2023,caiRealTimeBiddingReinforcement2017,yuReinforcementLearningHealthcare2023,nazariReinforcementLearningSolving2018}. 
The entity running the RL algorithm is called the
\emph{decision-maker}, which could be a dating application, 
an ad displaying platform or a scientist running a medical experiment.  
In the classical RL framework, an interaction proceeds as follows. The decision-maker observes the 
current \emph{state} which could be the profile of a user, 
the context of an ad or the health status of the subject and decides on an
\emph{action} to take based on the observed state which could 
be a partner recommendation, the display of an ad or the
choice of treatment to give to a subject. 
 Conditionally on the
state and action, the
decision-maker receives a random numerical
\emph{reward} which could be a like of the recommended person, a click on the
displayed ad or a numerical measure of the subject's health. 
The next state is a random function
of the current state and action. The goal of reinforcement learning is 
to optimize the \emph{policy} used by the decision-maker to choose actions from
previous observations so that the sum of rewards gathered over time is maximized.

In classical RL, the focus is on the decision-maker and the constraints
of the \emph{user}, which
could be a client of a dating application, a viewer of the ad, or a
subject of the medical experiment are not taken into account.
A primary concern of users in the above applications is privacy. 
A user may not want to publicly disclose their activity on a dating application,
their browsing history, or their health condition. 
While the decision-maker may be trustworthy, it can be hacked or leak data by
accident. For instance, the dating application Ashley Madison was hacked in 2015 
exposing the data of 32 million users~\cite{jonesHavingAffairMay2016}. Another example is 
the UK Biobank data containing health and genetic information of 500~000
subjects that was partially uploaded on Github by
accident~\cite{rocherProtectingHealthData2026}.  

While the UK Biobank data is de-identified, meaning the personal information such
as the name or the address of the subjects is removed, journalists from the
Guardian were able to re-identify some subjects~\cite{guardian}. The fact that 
removing personal information is not sufficient to protect the privacy of users
is well-documented~\cite{sweeney2000simple,narayananRobustDeanonymizationLarge2008,rocherEstimatingSuccessReidentifications2019} and justifies the use of more rigorous privacy frameworks such as
local differentially privacy (LDP)~\cite{duchiLocalPrivacyStatistical2013}.

The above considerations motivate the study of reinforcement learning under LDP.
We consider a tabular RL problem with $S$ states and $A$ possible actions, where the decision-maker learns an optimal policy by interacting with $K$ different users sequentially. 
The \emph{transition kernel}---the probabilities of moving from one state to another when an action is taken---and reward distributions are fixed (but unknown) and the same across all users.
Each user interaction corresponds to one \emph{episode}, where the initial state
varies from users to users.
When interacting with a user the decision-maker sends a policy to the user who
executes it for $H$ steps and applies a randomized \emph{mechanism} to privatize 
its data before sending the privatized data back to the decision-maker.
The mechanism is chosen by the decision-maker but must satisfy 
$\epsilon$-LDP constraints which guarantee a limit on the amount 
of information that the decision-maker - or hackers - can learn about the user.

\paragraph{Our contributions}  We first study the case $H=1$. We prove that the
minimax regret rate is given by 
\[
\min\bigg(K,  \frac{ \sqrt{\max( S, \exp(\epsilon) + 1) S A K (\exp(\epsilon) + 1)}}{\exp(\epsilon) - 1}\bigg)
\]
up to constants and a $\sqrt{\log(A)}$ factor. This result is a consequence
of~\Cref{thm:lower_bound} showing the lower bound and~\Cref{thm:upper_bound}
providing an algorithm and a mechanism whose regret matches the lower bound up
to constants and a $\sqrt{\log(A)}$ factor. 


Second, we study the case $H > 1$. A reduction from the bandit case implies a
new lower bound shown in~\Cref{cor:lower_bound} improving upon previous
work~\cite{garcelonLocalDifferentialPrivacy2021} uniformly across all
parameters. The improvement is particularly significant when $\epsilon \leq 1$,
gaining, in this case, a factor $\sqrt{S}$ over the previously known lower bound. 
Then, we present~\Cref{thm:upper_bound_rl} which provides an algorithm and 
a mechanism whose regret, for $\epsilon \leq 1$, is in $\tilde{O}( \frac{H^2 S^2\sqrt{ AK}}{\epsilon} )$ 
improving, in this high privacy regime, over the previous upper bound~\cite[Theorem
5]{garcelonLocalDifferentialPrivacy2021} by a factor of $H
\sqrt{A}$. We emphasize that, for $\epsilon \leq 1$, a multiplicative gap of
order $HS$ remains between this upper bound and the lower bound
of~\Cref{cor:lower_bound}, which is in $\Omega(\frac{HS\sqrt{AK}}{\epsilon})$.
Closing this gap is left as an open question (see~\Cref{sec:conclusion}).

\section{Related Work}
Regret minimization for non-private time-homogeneous episodic tabular RL is well
studied. A lower bound for the minimax regret of
order $\Omega(H \sqrt{ S A K})$ is obtained
by~\cite{azarMinimaxRegretBounds2017a} thanks to a reduction to contextual
bandits. Many works~\cite{azarMinimaxRegretBounds2017a,
dannPolicyCertificatesAccountable2019,
zanetteTighterProblemdependentRegret2019,zhangReinforcementLearningMore2021}
study regret minimization in RL and various algorithms have been proposed. 
The best upper bound up to date
is given by a model-based algorithm that achieves a regret in 
$\tilde{O}(H \sqrt{S A K} + H S^2 A)$ up to log factors~\cite{zhangReinforcementLearningMore2021}.
This result shows that tabular RL is essentially as hard as contextual bandits. 
Whether a model-free algorithm
can achieve a leading rate of $H \sqrt{SAK}$ is 
an active area of research~\cite{jinQlearningProvablyEfficient2018,
zhangAlmostOptimalModelFree2020}. 

The combination of online decision-making and local differential privacy is well
studied in the bandits case. In classical multi-armed bandits, the optimal
regret under $\epsilon$-LDP is in $\Theta(\max(\frac{\sqrt{AK}}{\epsilon},
\sqrt{AK}))$~\cite{renMultiArmedBanditsLocal2020}.  
We recover this rate up to constants and log factor by setting $S=1$ in
\Cref{thm:lower_bound} and \Cref{thm:upper_bound}. 

The works most related to ours are those studying RL under LDP constraints~\cite{garcelonLocalDifferentialPrivacy2021,qiaoNearoptimalDifferentiallyPrivate2023,chowdhuryDifferentiallyPrivateRegret2022}.
The work of~\cite{garcelonLocalDifferentialPrivacy2021} studies the same
time-homogeneous setting
as us, while~\cite{qiaoNearoptimalDifferentiallyPrivate2023}
and~\cite{chowdhuryDifferentiallyPrivateRegret2022} 
study the time-varying case where rewards and transition matrices change over
time. Since their bounds are for this more general setting, they are loose if applied directly. 
Our main point of comparison is therefore~\cite{garcelonLocalDifferentialPrivacy2021}.
We improve the lower bound obtained
in~\cite{garcelonLocalDifferentialPrivacy2021} and give an upper bound improving
that of~\cite{garcelonLocalDifferentialPrivacy2021}
in the high privacy regime $\epsilon \leq H\sqrt{A}$.

Although tabular contextual bandit under LDP has never been studied specifically, some generalizations 
were considered in the literature. Contextual linear bandits under LDP have been studied in~\cite{zheng2020locally}. After embedding the
reward using one-hot features and rescaling so that the unit-norm assumption on
the parameter is satisfied, the guarantee of
Zheng et al. gives in the tabular setting a rate in
$\tilde{O}\big(\frac{(SA)^{5/4}K^{3/4}}{\epsilon}\big)$, which is suboptimal.
If contexts are chosen i.i.d., then, under some assumptions, an
$O(K^{1/2})$ rate can be recovered~\cite{hanGeneralizedLinearBandits2021,pavlovic2025differential}.
The setting we study differs in that there are no restrictions on how contexts
are chosen. Another generalization is contextual nonparametric bandits under
LDP, which have been studied in~\cite{ma2025locally},
where the authors obtain sublinear regret under regularity assumptions
(Assumptions 2.2 and 2.3) and a margin assumption (Assumption 2.4). These
assumptions do not necessarily hold in the setting we study.
Lastly, a related problem is LDP contextual-bandit pricing~\cite{zhao2026contextual,chen2023differential},
where the price is the action and the revenue is the reward. However, none of
the results of~\cite{zhao2026contextual} or~\cite{chen2023differential} apply in the setting we consider
because of incompatible assumptions. Notably, in Zhao et al., the reward has a
specific parametric form derived from a GLM assumption, and in Chen et al., the
reward is assumed to be strongly concave. In contrast, none of these
restrictions are imposed in the setting we study.

Besides LDP, other notions of privacy have been studied in bandits and RL.
In \emph{global differential privacy}, the decision-maker is trusted but the sequence of action it takes
must not reveal too much information about the users'
reward~\cite{azizeWhenPrivacyMeets2022,azizeDifferentiallyPrivateBestArm2026}.
This setting is only relevant for multi-armed bandits since in contextual
bandits and RL, a linear regret is unavoidable~\cite{shariffDifferentiallyPrivateContextual2018}.
This motivates the notion of \emph{joint differential
privacy}~\cite{shariffDifferentiallyPrivateContextual2018,vietriPrivateReinforcementLearning2020} 
which is weaker. It guarantees that the actions taken by the decision-maker at time
$t$ do not reveal too much information about the users' information up to time
$t'<t$. Intermediate notions based on shuffling have also been studied for
linear contextual bandits~\cite{chowdhury2022shuffle,garcelon2022privacy}.





\section{Episodic Finite-Horizon Tabular RL under LDP}
\label{sec:setting}
\paragraph{Notations} For an integer $n$, we use $[n]$ to denote the set $\{1, 2, \ldots, n\}$.
For a finite set $\mathcal{X}$, we use $\Delta(\mathcal{X})$ to denote the set
of functions $\nu: \mathcal{X} \rightarrow [0, 1]$ such that $\sum_{x \in \Xcal}
\nu(x) = 1$ and denote by $\nu(\cdot)$ the corresponding distribution over $\Xcal$. 
We use $f(x) = \Omega(g(x))$ to denote that there exists a constant $C > 0$ such
that $f(x) \geq C g(x)$, $f(x) = O(g(x))$ to
denote that there exists a constant $C > 0$ such that $f(x) \leq C g(x)$ and
$f(x)= \Theta(g(x))$ if $f(x) = \Omega(g(x))$ and $f(x) = O(g(x))$. We use
$\tilde{O}(\cdot)$, $\tilde{\Omega}(\cdot)$ and $\tilde{\Theta}(\cdot)$ to hide constants and logarithmic
factors. We denote $\Bcal(p)$ the Bernoulli distribution with parameter $p$.
$D_\text{KL}(\nu, \nu') = \EE_{\nu} \left[\log \frac{d\nu}{d\nu'}\right]$ and 
$D_{\text{TV}}(\nu, \nu') = \sup_{A} |\nu(A) - \nu'(A)|$
denote the Kullback-Leibler divergence and total variation distance between 
distributions $\nu$ and $\nu'$.
\paragraph{Non-private interaction, value functions and regret}
A Markov Decision Process (MDP) is defined by the tuple
$I = (\Scal = [S], \Acal = [A], P, r, (\rho^k)_{k=1}^K)$, 
where $\Scal$ is the set of $S$ states, $\Acal = [A]$ the set of $A$ actions,
$P: \Scal \times \Acal \rightarrow \Delta(\Scal)$ 
is the \emph{transition kernel}, $r:\Scal \times \Acal \rightarrow [0, 1]$ is
the expected \emph{reward function} and $\rho^k \in \Delta(\Scal)$ is the
distribution 
of the starting state at episode $k$ for each $k \in [K]$.
We call $\mathfrak{M}(S, A, K)$, or simply $\mathfrak{M}$ when the context is clear, the set of MDP with $S$ states,
$A$ actions and $K$ episodes.
The above description corresponds to the \emph{tabular} case where the number of 
states, actions and episodes are finite. 

In the non-private setting, the decision-maker interacts directly with the MDP.
The interaction consists of $K$ episodes each consisting of $H$ steps. 
At episode $k$, the decision maker selects a policy $\pi^k = (\pi^k_h: \Scal
\rightarrow  \Delta(\Acal))_{h \in [H]}$ and samples a starting state $S^k_1$ from
$\rho^k$. 
Then, at step $h \in [H]$, the decision maker observes $S^k_h \in \Scal$, samples
an action $A^k_h \in \Acal$ from $\pi^k_h(S^k_h, \cdot)$,
obtains a reward $R^k_h \in [0, 1]$ and transitions to $S^k_{h+1} \in \Scal$.
The reward $R^k_h$ and the next state $S^k_{h+1}$ are
sampled independently conditionally on $(S^k_h, A^k_h)$, 
with $\EE[R^k_h \mid S^k_h, A^k_h] = r(S^k_h, A^k_h)$ and
$S^k_{h+1} \mid S^k_h, A^k_h \sim P(S^k_h, A^k_h, \cdot)$.
The \emph{trajectory} of episode $k$ is denoted by $\tau^k = (S_1^k, A_1^k, R_1^k,
\ldots, S_H^k, A_H^k, R_H^k, S_{H+1}^k) \in \mathbb{T}
$ where $\mathbb{T} = (\Scal \times \Acal \times [0, 1])^H \times \Scal$ is
the set of possible trajectories.
The \emph{value} of a policy $\pi$ at step $h$ on MDP $I$ at state $s$ is defined as 
\[
V^{I, \pi}_h(s) = \EE\left[\sum_{i=h}^H R^k_i \mid S_h^k = s \right].
\]
where the expectation is taken over the trajectory $\tau^k$. We often omit the 
superscript $I$ when it is clear from the context.
The \emph{optimal value function} is defined as $V^*_h(s) = \max_{\pi} V^{I, \pi}_h(s)$.
Lastly, note that the policies we use depend only on the current state and step. It can be shown
that using history-dependent policies instead 
does not increase the optimal value function~\cite[Theorem
4.4.2]{putermanMarkovDecisionProcesses2014}.
We call the process by which the decision maker updates policies from episode to episode 
the (non-private) \emph{algorithm} $\mathfrak{A} = (\mathfrak{A}_k: \mathbb{T}^{k-1}
\rightarrow (\Scal \rightarrow  \Delta(\Acal))^H)_{k \in [K]}$ and call
$\mathbb{A}^{\text{public}}$ the set of non-private algorithms.
where
$\mathfrak{A}_k$ maps the past $k-1$
trajectories $(\tau^1, \ldots, \tau^{k-1})$ to the policy $\pi^k$.

The performance of an algorithm $\mathfrak{A}$ on an MDP $I$ is computed by the \emph{regret} defined as
\[
\Rcal^{\text{public}}(\mathfrak{A}, I) = \EE\left[\sum_{k=1}^K \left( V^{I, *}_1(S_1^k) - V^{I, \pi^k}_1(S_1^k) \right)\right], \quad \pi^k \sim \mathfrak{A}_k(\tau^1, \ldots, \tau^{k-1}),
\]
and the difficulty of the problem is quantified by the \emph{minimax regret} defined as
\[
\mathfrak{R}^{\text{public}} = \inf_{\mathfrak{A} \in \mathbb{A}^{\text{public}}} \sup_{I \in \mathfrak{M}} \Rcal(\mathfrak{A}, I).
\]

\paragraph{Private interaction and local differential privacy} In the setting
we study, the user does not allow the decision maker to access its true data.
The algorithm will therefore not have access to the trajectories but only to a
version privatized by a mechanism $\mathcal{M}: \mathbb{T} \rightarrow \Delta(\Ycal)$ satisfying the following 
$\epsilon$-LDP guarantee: for any two trajectories $\tau, \tau'$ 
and any measurable subset $Y \subset \Ycal$, we must have
\begin{align*}
    \PP(\mathcal{M}(\tau) \in Y) \leq e^{\epsilon} \PP(\mathcal{M}(\tau') \in Y).
\end{align*}
We call $\mathbb{M}_\epsilon$ the set of such $\epsilon$-LDP mechanisms.

The interaction then proceeds as follows. The decision maker first selects
a privacy mechanism $\mathcal{M}$. 
Then, at episode $k$, the decision maker selects a policy $\pi^k$ and sends it to user
$k$, who executes the policy $\pi^k$ for $H$ steps and observes the resulting
trajectory $\tau^k$ which is then privatized by the mechanism $\mathcal{M}$ to
produce a privatized statistic $M^k = \mathcal{M}(\tau^k)$ sent back to the
decision maker. 

The process by which the decision maker updates policies from episode to episode
is now characterized by a \emph{private algorithm} $\mathfrak{A} =
(\mathfrak{A}_k: \Ycal^{k-1} \rightarrow (\Scal \rightarrow  \Delta(\Acal))^H)_{k
\in [K]}$ and the set of private algorithms is denoted by $\mathbb{A}^{\text{private}}$. 
Here $\Ycal^{k-1}$ is the Cartesian-product space containing all possible
sequences of $k-1$ private messages: at episode $k$, the realized input to the
algorithm is the history $(M^1,\ldots,M^{k-1}) \in \Ycal^{k-1}$.

Furthermore, the performance of the decision maker is now determined by the choice of mechanism
$\mathcal{M}$ and the private algorithm $\mathfrak{A}$. We still measure the
performance by the regret given by 
\begin{align*}
\Rcal(\mathfrak{A}, I, \mathcal{M}) = \EE\left[\sum_{k=1}^K \left( V^{I, *}_1(S_1^k) - V^{I, \pi^k}_1(S_1^k) \right)\right], \quad \begin{aligned} M_1, \dots, M_K \sim \mathcal{M}(\tau^1), \dots, \mathcal{M}(\tau^K) \\
\pi^k \sim \mathfrak{A}_k((M_i)_{i \in [k-1]})\end{aligned}
\end{align*}
and the minimax regret is defined as
\[
\mathfrak{R} = \inf_{\substack{\mathfrak{A} \in \mathbb{A}^{\text{private}} \\ \mathcal{M} \in \mathbb{M}_\epsilon}} \sup_{I \in \mathfrak{M}} \Rcal(\mathfrak{A}, I, \mathcal{M}).
\]

In the following sections, we present upper and lower bounds on $\mathfrak{R}$.

\section{Private contextual bandits}
The current lower bounds available in the literature in 
either the non-private~\cite{azarMinimaxRegretBounds2017a} or
 private~\cite{garcelonLocalDifferentialPrivacy2021} case use
 a reduction to contextual bandits. It is therefore natural to first
 study this simpler case.
We therefore set in this section $H=1$. In this case, the transition
kernel $P$ plays no role and since $A$ and $S$ are fixed, an instance of
contextual bandits is characterized by the
reward function $r$ and the initial distributions $\{\rho^k\}_{k=1}^K $.

\subsection{Lower bound}
In~\Cref{thm:lower_bound}, we show a lower bound on $\mathfrak{R}$.
\begin{theorem}
  \label{thm:lower_bound}
    Assume $A \geq 2$, for any $\epsilon$-LDP mechanism $\mathcal{M}$ and any algorithm
    $\mathfrak{A}$ that the decision maker can choose, there exists an instance $I$ characterized by a
    reward function $r$ and initial distributions $\{\rho^k\}_{k=1}^K$ such that
    the regret is lower bounded as
    \begin{align*}
        \Rcal(\mathfrak{A}, \mathcal{M}, I) \geq \frac1{32} \min\bigg(K,  \frac{ \sqrt{\max( S, \exp(\epsilon) + 1) S A K (\exp(\epsilon) + 1)}}{\exp(\epsilon) - 1}\bigg)
    \end{align*}
\end{theorem}

\begin{proof}[Proof sketch (see \Cref{app:lower_bound} for the full proof)]
Fix an algorithm $\mathfrak{A}$, a mechanism $\mathcal{M}$ and a gap $\Delta>0$. For each $\phi\in\Acal^{\Scal}$ and $\sigma\in\Scal^K$, define the hard instance $I_{\phi,\sigma}$ by
\[
r_{\phi,\sigma}(s,a)=\frac12+\Delta\ind{a=\phi(s)},\qquad \rho^k_{\phi,\sigma}(s)=\ind{s=\sigma_k}.
\]
Let $I_{\alpha,c}$ be sampled uniformly from this family and let $I_s^0$ be the same random instance with all rewards at state $s$ replaced by $1/2$. A standard change-of-measure argument lower bounds the expected regret by the gap times the probability of not identifying the best action, and upper bounds this identification probability through $D_{\rm KL}((M_1,\ldots,M_K)_{s,0}\|(M_1,\ldots,M_K)_{s,j})$.

The improvement comes from controlling this KL term using both LDP contractions of~\cite[Theorems 1 and 2]{asoodehContractionLocallyDifferentially2024}, namely
\begin{align*}
D_{\rm KL}(M_1\|M_2)&\leq \frac{(\exp(\epsilon)-1)^2}{(\exp(\epsilon)+1)^2}D_{\rm KL}(\tau_1\|\tau_2),\\
D_{\rm KL}(M_1\|M_2)&\leq \frac{(\exp(\epsilon)-1)^2}{\exp(\epsilon)}D_{\rm TV}(\tau_1,\tau_2)^2.
\end{align*}
Combining these inequalities with the chain rule yields the two regimes in the bound. Optimizing $\Delta$ and using the trivial upper bound $\Rcal\leq K$ concludes.
\end{proof}

When $\epsilon \leq 1$, the lower bound of~\Cref{thm:lower_bound} is in
$\Omega(\frac{S \sqrt{A K}}{\epsilon})$. Compared to the non-private rate of
$\tilde{\Theta}(\sqrt{S A K})$ obtained by~\cite{azarMinimaxRegretBounds2017a}, there is
a multiplicative cost of $\frac{\sqrt{S}}{\epsilon}$ due to privacy. 
When $\epsilon \leq 1$, the rate of private bandits with $SA$ arms is in $\Theta(\sqrt{SAK}/\epsilon)$, the results of~\Cref{thm:lower_bound} shows 
that private contextual bandit with $S$ context and $A$ arms is strictly harder as a factor of order $\sqrt{S}$ is gained.

Intuitively the difficulty in private bandits with $S$ context and $A$ arms is that since context must remain private, 
when a privatized reward is received it is not clear from which bandit
it came. The computation of the TV distance in the proof of~\Cref{thm:lower_bound} in~\Cref{app:lower_bound} shows that a factor $\frac1{S}$ appears
become of the randomness of the context. This ultimately leads to the lower bound in $\frac{S \sqrt{AK}}{\epsilon}$ 
in the high privacy regime $\epsilon \leq 1$.

\begin{remark}[Sequentially interactive mechanisms]
\label{rem:sequentially_interactive}
In the setting of~\Cref{sec:setting}, the same mechanism $\mathcal{M}$ privatizes
every trajectory. \Cref{thm:lower_bound} holds more generally for
\emph{sequentially interactive} $\epsilon$-LDP mechanisms
(\Cref{def:sequentially_interactive} in~\Cref{app:lower_bound}), for which the
$\epsilon$-LDP channel used at episode $k$ may depend on the previous messages
$M^1, \ldots, M^{k-1}$. The proof only uses that, given the past messages, the
same $\epsilon$-LDP channel is applied under the two instances being compared,
whether or not this channel depends on the history. The proof
in~\Cref{app:lower_bound} is written for this more general class.
\end{remark}

\subsection{Upper bound}
In this subsection, we show that the lower bound of~\Cref{thm:lower_bound} is
essentially tight by showing that there exists an algorithm and a mechanism
achieving a regret upper bound matching the lower bound up to constants and a
multiplicative $\sqrt{\log(A)}$ factor.

The algorithm we use is based on Exp3 with forced
exploration~\cite{auerNonstochasticMultiarmedBandit2002} 
and the mechanism we use is based on the optimized unary encoding
mechanism~\cite{wangLocallyDifferentiallyPrivate2017}. 
We name the procedure Exp3-OUE and give an implementation in~\Cref{alg:exp3}.

We first describe the optimized unary encoding mechanism (OUE). Given
$v \in \{0, 1\}^d$ with $\|v\|_1 \leq 1$, the OUE mechanism outputs a vector $\text{OUE}(v) \in \{0, 1\}^d$ 
where each coordinate is generated independently as follows: 
\begin{align}
&\text{OUE}: v \in \{w \in \{0, 1\}^d: \|w\|_1 \leq 1\} \rightarrow \text{OUE}(v) \in \{0, 1\}^d \quad \text{such that} \notag \\
&\forall i \in [d], \text{OUE}(v)_i \mid v_i \sim
\begin{cases} 
  \Bcal(\frac12), \text{if} ~ v_i = 1 \\
  \Bcal\bigg(\frac{1}{\exp(\epsilon) + 1}\bigg) , \text{if} ~ v_i = 0
\end{cases}. \label{eq:OUE}
\end{align}
As shown in~\cite{wangLocallyDifferentiallyPrivate2017}, OUE is $\epsilon$-LDP. 
This mechanism is optimal for histogram estimation under LDP~\cite{canonneLocallyPrivateHistograms2024}. It is therefore a natural
candidate for our problem.

We now describe Exp3-OUE informally. At episode 
$k$, the user receives a deterministic policy $\pi^k$ from the decision-maker and executes it,
which yields a trajectory $\tau^k = (S^k, A^k, R^k)$. Then, the user binarizes
the reward by sampling $B^k \sim \Bcal(R^k)$ and constructs a vector $V^k$
of dimension $S$ where $V^k(s) = \ind{S^k=s, B^k=1}$.
This vector is then privatized by OUE yielding a
vector $M^k = \text{OUE}(V^k)$ sent back to the decision-maker. 
The decision maker then builds an importance-weighted estimator $\hat V^k(s, a)$ of the reward
for each state $s \in \Scal$ and action $a \in \Acal$ and uses it to update the
weights of Exp3 as $w_{k+1} = w_k \exp(\eta \hat V^k)$ where $\eta$ is a
learning rate parameter and $w_k$ is known from previous episodes (we
set $w_1 = (1, \dots, 1)$). Following the Exp3 algorithm, these weights are used to
compute the new policy as in \Cref{eq:policy_update}.
For every state $s \in \Scal$, we sample $\alpha_s$ from $\Pi^{k+1}(\cdot, s)$
and define the deterministic policy $\pi^{k+1}$ by $\pi^{k+1}(s) = \alpha_s$.
The policy $\pi^{k+1}$ is then sent to the user of episode $k+1$ and the process
continues.

\begin{algorithm}[hbtp]
  \caption{Exp3 with optimized unary encoding for private contextual bandits (Exp3-OUE)}
	\label{alg:exp3}
	\SetKwInput{KwData}{Input}
\SetKwComment{Comment}{\color{purple} $\triangleright$\ }{}
 	\KwData{$\epsilon$, $K$, $\Acal$, $\Scal$, $\eta$, $\gamma$}

  \Comment{\color{purple} Decision maker (public operations)}

  Initialize $w_1(s, a) = 1$ for all $s \in \Scal$ and $a \in \Acal$

  Define the distributions $\Pi^1(\cdot, s)$ as $\Pi^1(a, s) = (1 -
  \gamma)\frac{w_1(s, a)}{\sum_{a' \in \Acal} w_1(s, a')} + \frac{\gamma}{A}$ for
  each $s, a \in \Scal \times \Acal$ 

  Set $\pi^1(s) = \alpha_s$ where $\alpha_s \sim \Pi^1(\cdot, s)$ and send $\pi^1$ to user $1$.

  \For{$k = 1$ \KwTo $K$}{
    \Comment{\color{purple} User $k$ (private operations)}

    Observe $S^k$, select $A^k = \pi^k(S^k)$, observe $R^k$, and sample $B^k
    \sim \Bcal(R^k)$

    Construct $V^k(s) = \ind{S^k=s} B^k$ for all $s \in \Scal$

    Construct $M^k \in \{0, 1\}^{\Scal} = \text{OUE}(V^k)$ via the OUE
    mechanism~\Cref{eq:OUE}

    Send $M^k$ to the decision maker

    \Comment{\color{purple} Decision maker (public operations)}
    For each $s \in \Scal$ and $a \in \Acal$, construct the estimator
    \begin{align*}
        \hat{V}^k(s, a) = M^k(s) \frac{\ind{a = \pi^k(s)}}{\Pi^k(a, s)}
    \end{align*}

    Update the weights as for each $s, a \in \Scal \times \Acal$,
    \begin{equation*}
    w_{k+1}(s, a) = w_k(s, a) \exp(\eta \hat{V}^k(s, a)).
    \end{equation*}

    Define $\Pi^{k+1}(\cdot, s)$ as for each $s, a \in \Scal \times \Acal$,
    \begin{equation}
      \label{eq:policy_update}
    \Pi^{k+1}(a, s) =
    (1 - \gamma)\frac{w_{k+1}(s, a)}{\sum_{a' \in \Acal} w_{k+1}(s, a')} + \frac{\gamma}{A}.
    \end{equation}

    Set $\pi^{k+1}(s) = \alpha_s$ where $\alpha_s \sim \Pi^{k+1}(\cdot, s)$ for
    each $s \in \Scal$ and send $\pi^{k+1}$ to user $k+1$.
  }
\end{algorithm}

In essence, Exp3-OUE is Exp3 applied to privatized input, but two implementation
choices make a significant difference compared to a direct Laplace-based
LDP-OBI~\cite{garcelonLocalDifferentialPrivacy2021} adapted to the bandit case.
First, in Exp3-OUE, the decision-maker
sends a
deterministic
realization $\pi^k$ of $\Pi^k$ rather than $\Pi^k$ directly, so the user privatizes a vector indexed by states rather than by state-action pairs, improving the dependence on $A$.
Second, the user binarizes the reward and applies OUE, which improves the
dependence on $\epsilon$. The estimator $\hat V^k(s,a)$ also gives the right
bias-variance trade-off; debiasing it would increase the regret.

In~\Cref{thm:upper_bound}, we precisely characterize the regret of
Exp3-OUE and show that it matches the lower bound of~\Cref{thm:lower_bound} 
up to constants and a multiplicative $\sqrt{\log(A)}$ factor.

\begin{theorem}
\label{thm:upper_bound}
    Assume $A \geq 2$, for any instance $I$ characterized by a reward function
    $r$ and initial distributions $\{\rho^k\}_{k=1}^K$, the regret of
    \Cref{alg:exp3} run with $\gamma = A \eta$ and \[ \eta = \min(\frac1{2A}, \sqrt{\frac{S \ln(A)}{2(e-1) K A \big(\frac12 - \frac1{\exp(\epsilon) + 1} + \frac{S}{\exp(\epsilon) + 1}\big)}}) \] is upper bounded as
    \begin{align*}
        \Rcal(\mathfrak{A}, \mathcal{M}, I) \leq 16 \min\bigg(K,  \frac{ \sqrt{\max( S, \exp(\epsilon) + 1) S A K  (\exp(\epsilon) + 1) \log(A)}}{\exp(\epsilon) - 1}\bigg)
    \end{align*}
\end{theorem}

\begin{proof}[Proof sketch (see \Cref{app:upper_bound} for the full proof)]
The proof follows the standard Exp3 potential argument~\cite[Theorem 3.1]{auerNonstochasticMultiarmedBandit2002}. Let
\[
\Phi_k=\sum_{s\in\Scal}\log\Big(\sum_{a\in\Acal}w_k(s,a)\Big).
\]
Since $\hat V^k(s,a)\in[0,A/\gamma]$ and $\gamma=A\eta$, we have $\eta\hat V^k(s,a)\in[0,1]$. Expanding $\Phi_{k+1}-\Phi_k$, summing over $k$, using $\Phi_1=S\log(A)$ and comparing with the best fixed action $a^*(s)$ gives, after taking expectations,
\begin{align*}
\EE\Big[\sum_{k=1}^K r(S^k,a^*(S^k))-\sum_{k=1}^K r(S^k,A^k)\Big]
&\leq \frac{S\ln(A)}{\eta(\frac12-\frac1{\exp(\epsilon)+1})}\\
&\quad +\frac{2(e-1)KA\eta(\frac12-\frac1{\exp(\epsilon)+1}+\frac{S}{\exp(\epsilon)+1})}{\frac12-\frac1{\exp(\epsilon)+1}}.
\end{align*}
Plugging in the chosen learning rate and using that regret is at most $K$ yields the theorem.
\end{proof}

As a consequence of \Cref{thm:lower_bound} and \Cref{thm:upper_bound}, 
we have that the lower bound in~\Cref{thm:lower_bound} is tight up to
log-factor. Therefore, the minimax regret $\mathfrak{R}$ for private contextual bandits satisfies
\[
\mathfrak{R} = \tilde{\Theta}\bigg( \min\bigg(K,  \frac{ \sqrt{\max( S, \exp(\epsilon) + 1) S A K (\exp(\epsilon) + 1)}}{\exp(\epsilon) - 1}\bigg)\bigg).
\]


\section{Private Tabular Reinforcement Learning}
In this section, we attempt to extend the lower bound of \Cref{thm:lower_bound}
and the upper bound of \Cref{thm:upper_bound} to the case of finite-horizon
MDPs.
We first focus on lower bound and see in~\Cref{cor:lower_bound} that the usual reduction from
reinforcement learning to contextual bandits applies in the private setting as
well. This yield a lower bound for private tabular reinforcement learning that improves over
previous work~\cite{garcelonLocalDifferentialPrivacy2021}.
We then establish an upper bound in~\Cref{thm:upper_bound_rl}, adapting the
ideas used to prove~\Cref{thm:upper_bound} to the case of finite-horizon MDPs.
In the high privacy regime $\epsilon\leq 1$, this upper bound improves
over known upper bounds for private tabular reinforcement 
learning~\cite{garcelonLocalDifferentialPrivacy2021,
qiaoNearoptimalDifferentiallyPrivate2023,
chowdhuryDifferentiallyPrivateRegret2022}.

\subsection{Lower bound}
We first show the following corollary of \Cref{thm:lower_bound}.
\begin{corollary}
  \label{cor:lower_bound}
    Assume $S \geq 4$ and $A \geq 2$ and $H \geq 2$. For any $\epsilon$-LDP mechanism $\mathcal{M}$ and any algorithm
    $\mathfrak{A}$ that the decision maker can choose, there exists an instance $I$ of finite-horizon MDP such that
    the regret is lower bounded as
    \begin{align*}
        \Rcal(\mathfrak{A}, \mathcal{M}, I) \geq \frac{H}{128} \min\bigg(K,  \frac{ \sqrt{\max( S, \exp(\epsilon) + 1) S A K (\exp(\epsilon) + 1)}}{\exp(\epsilon) - 1}\bigg)
    \end{align*}
\end{corollary}
\begin{proof}[Proof sketch (see \Cref{app:cor_lower_bound} for the full proof)]
  Consider a finite-horizon MDP with $\Scal = [S], \Acal = [A]$. 
  $\rho^k$ has support over $[S-2]$. The other 2 states are labelled "good" and "bad". 
  Both good and bad states are absorbing and yield a reward of $1$ and $0$
  respectively.
  For any $s \in [S-2]$, any actions lead to the good state with some
  probability and to the bad state with the remaining probability with the
  reward being $0$ in both cases. 
  This MDP is identical to a contextual bandit problem with $S-2$ contexts and $A$
  actions except that rewards are scaled by $H-1$.
  Then, choose $P$ and $\rho$ such that the hard instance
  of~\Cref{thm:lower_bound} is obtained and the lower bound is proved.
\end{proof}

The lower bound in~\cite{garcelonLocalDifferentialPrivacy2021} is given (up to constants) by 
$
 H \frac{\sqrt{SAK}}{\min(\exp(\epsilon) - 1, 1)}
$ a rate that is lower than that of~\Cref{cor:lower_bound} for any value of $H, S, A, K, \epsilon$.
More precisely, the two bounds coincide when $\epsilon$ is on the order of $\log S$ or larger, and when $\epsilon \leq \log S$ the improvement is a
multiplicative factor of order $\sqrt{S/\exp(\epsilon)}$.

The reason why the lower bound in~\Cref{thm:lower_bound} dominates is because in~\cite{garcelonLocalDifferentialPrivacy2021},
authors construct a tree  in which one starts at the root and moves deterministically depending on
the actions played, the number of leaves being of order $S$, and in which one
special leaf-action pair is slightly more rewarding than the others. Their
construction therefore reduces the problem to a private bandit instance with
$SA$ arms. In contrast, our lower bound reduces the problem to a private contextual bandit with $S$ context and $A$ arms 
which is a more difficult problem.

\subsection{Upper bound}
In this sub-section, we show an upper bound for the minimax regret improving 
over known upper bounds in the high privacy regime.It is obtained using a variant of UCBVI as the algorithm 
and the optimized unary encoding as the
$\epsilon$-LDP mechanism. 

We name the procedure UCBVI-OUE, which we now present
informally. A precise implementation is available in~\Cref{alg:ucbvi}.
At episode $k$, the user receives a deterministic policy $\pi^k$ and an index $J^k$ from the
decision-maker.
The user executes the policy $\pi^k$, which yields a trajectory $\tau^k = (S^k_1, A^k_1, R^k_1, \dots,
S^k_H, A^k_H, R^k_H)$. 
Then, the user binarizes the
reward at step $J^k$ by $B^k_{J^k} \sim \Bcal(R^k_{J^k})$.
The user then constructs a vector $I^k = (\ind{S^k_{J^k}=s, B^k_{J^k} = b,
S^k_{J^k+1}=s'})_{s \in \Scal, b \in \{0, 1\}, s' \in \Scal}$ 
of dimension $2 S^2$ and privatizes it using the OUE mechanism to get $M^k =
\text{OUE}(I^k)$.
$M^k$ is then sent back to the decision-maker who 
uses it to build for each $s \in \Scal$ and $a \in \Acal$, confidence sets $\Ccal^k(s, a)$ which contain $P(s, a, \cdot)$
and $r(s, a)$ with high probability.
Then, the decision maker computes optimistic $Q$-value estimates $\tilde{Q}^{k,
(+)}$ using the confidence sets by backward induction as in
\Cref{eq:optimistic_q}. Then the new policy is computed as $\pi^{k+1}_h(s) =
\argmax_{a \in \Acal}
\tilde{Q}^{k, (+)}_h(s, a)$ for each $s \in \Scal$ and $h \in [H]$ and $J^{k+1}$
is sampled uniformly at random from $[H]$. $\pi^{k+1}$ and $J^{k+1}$ are then sent to the
user of episode $k+1$ and the process continues.

In addition to the two ideas already used in the case of contextual bandits, 
which are sending a deterministic policy instead of a randomized one and
binarizing the reward to use the OUE mechanism, UCBVI-OUE also uses the idea of
sending the indicator of only a triplet $S^k_{J^k},
B^k_{J^k}, S^k_{J^k+1}$ from the trajectory  
chosen uniformly at random as opposed to sending the indicator of the whole
trajectory. This reduces the amount of noise that needs to be added to ensure LDP 
and ultimately leads to an estimator that has lower variance.

\begin{algorithm}
  \caption{UCBVI with optimized unary encoding for private tabular RL (UCBVI-OUE)}
  \label{alg:ucbvi}
  \SetKwInput{KwData}{Input}
\SetKwComment{Comment}{\color{purple} $\triangleright$\ }{}
\KwData{$\epsilon$, $K$, $\Acal$, $\Scal$, $H$, $\Upsilon_\epsilon =
\frac{\exp(\epsilon) + 1}{\exp(\epsilon) - 1}$, $q_\epsilon =
\frac{1}{\exp(\epsilon) + 1}$.
}
  \Comment{\color{purple} Decision maker (public operations)}
Initialize $\tilde{Q}^0_h(s, a) = H$ for all $s,a,h \in \Scal \times \Acal \times [H]$ 

  Send $\pi^1_h(s) = \argmax_{a \in \Acal} \tilde{Q}^0_h(s, a)$ to user $1$ for each $s \in \Scal$ and $h \in [H]$.

  Sample $J^1$ uniformly at random from $[H]$ and send $J^1$ to user $1$.

  \For{$k = 1$ \KwTo $K$}{
    \Comment{\color{purple} User $k$ (private operations)}

    Play according to policy $\pi^k$ and observe $(S^k_h, A^k_h, R^k_h,
    S^k_{h+1})_{h \in [H]}$

    Sample $B^k_h \sim \Bcal(R^k_h)$ for each $h \in [H]$

    Construct $I^k(s, b, s') = \ind{S^k_{J^k}=s, B^k_{J^k} = b, S^k_{J^k+1}=s'}$ for
    all $s, s' \in \Scal$, $b \in \{0, 1\}$. 

    Construct $M^k \in \{0, 1\}^{\Scal} = \text{OUE}(I^k)$ via the OUE
    mechanism~\Cref{eq:OUE}

    Send $M^k$ to the decision maker

  \Comment{\color{purple} Decision maker (public operations)}
  For each $s, b, s' \in \Scal \times \{0, 1\} \times \Scal$, define $\hat{I}^k(s, b, s') = 2
  \Upsilon_\epsilon (M^k(s, b, s') - q_\epsilon)$ 

  For each $s, a, b, s' \in \Scal \times \Acal \times \{0, 1\} \times \Scal$,
  compute $\beta(s, a)$ as in~\Cref{eq:beta_k} and define
  \begin{align}
  &\hat{N}^k(s, a, b, s') = H \sum_{i \in [k]}  \hat{I}^i(s, b, s') \ind{\pi^i_{J^i}(s) = a} \label{eq:count} \\
  & \tilde{N}^{k, (-)}(s, a) =  \max(1, \sum_{b \in \{0, 1\}, s' \in \Scal}  (\hat{N}^k(s, a, b, s') - \beta_k(s, a)))  \label{eq:estimator_count}
  \end{align}
  Then, build confidence sets 
\begin{align}
  \label{eq:confidence_set}
  \Ccal^k(s, a) = \bigg\{\nu \in \Delta(\Scal), &\rho \in [0, 1]: |\tilde{N}^{k, (-)}(s, a) \rho - \sum_{s' \in \Scal} \hat{N}^k(s, a, 1, s') | \leq 7 S \beta_k(s, a) \notag \\
  & \|\tilde{N}^{k, (-)}(s, a) \nu(\cdot) - \sum_{b \in \{0, 1 \}} \hat{N}^k(s, a, b, \cdot) \|_1 \leq 8 S\beta_k(s, a)  \bigg\}
\end{align}
Set $\tilde{Q}^{k, (+)}_{H+1}(s, a) = 0$ for each $s, a \in \Scal \times \Acal$ and then compute
\begin{equation}
  \label{eq:optimistic_q}
\tilde{Q}^{k, (+)}_h(s, a) = \max_{\nu, \rho \in \Ccal^k(s, a)} \big(\rho + \sum_{s' \in \Scal}
\nu(s') \max_{a' \in \Acal} \tilde{Q}^{k, (+)}_{h+1}(s', a') \big)
\end{equation}

  Send $\pi^{k+1}_h(s) = \argmax_{a \in \Acal} \tilde{Q}^{k, (+)}_h(s, a)$ to user $k+1$ for each $s \in \Scal$ and $h \in [H]$.

  Sample $J^{k+1}$ uniformly at random from $[H]$ and send $J^{k+1}$ to user $k+1$.
  }
\end{algorithm}

We now present \Cref{thm:upper_bound_rl}, which gives an upper bound on the regret of \Cref{alg:ucbvi}.
\begin{theorem}
\label{thm:upper_bound_rl}
     Assume $S \geq 4$, $A \geq 2$ and $H \geq 2$.
    For any instance $I$ of finite-horizon MDP, the regret of \Cref{alg:ucbvi}
    is upper bounded up to constants and log-factors (see the exact expression
    in~\Cref{app:upper_bound_rl}) as
    \begin{align*}
        \Rcal(\mathfrak{A}, \mathcal{M}, I) \leq &\tilde{O} \bigg(H^2S^2 \sqrt{ \Upsilon_\epsilon^2 AK } + H^2S^2 A \Upsilon_\epsilon  \bigg), \quad \Upsilon_\epsilon = \frac{\exp(\epsilon) + 1}{\exp(\epsilon) - 1}
    \end{align*}
\end{theorem}

\begin{proof}[Proof sketch (see \Cref{app:upper_bound_rl} for the full proof)]
Let
\[
N^k(s,a,b,s')=\sum_{i\in[k]}\sum_{h=1}^H\ind{S^i_h=s,B^i_h=b,S^i_{h+1}=s',\pi^i_h(s)=a}
\]
and let $\hat N^k$ be the private estimator in~\Cref{eq:estimator_count}. Freedman's inequality~\cite{hsuMixingTimeEstimation2015} implies (\Cref{lemma:deviation_hatN}) that, with probability at least $1-\delta$, for all $s,a,b,s',k$,
\begin{align}
&|\hat{N}^k(s,a,b,s')-N^k(s,a,b,s')|\leq \beta_k(s,a), \notag \\
&\beta_k(s,a)=\sqrt{20H\Upsilon_\epsilon^2\sum_{h\in[H],i\in[k]}\ind{a=\pi^i_h(s)}\log\frac{48K^2S^2A}{\delta}} + 4H\Upsilon_\epsilon\log\frac{48K^2S^2A}{\delta}. \label{eq:beta_k}
\end{align}
Furthermore, \Cref{lemma:properties_confidence_set} shows $\Ccal^k(s, a)$
\Cref{eq:confidence_set} contains with high probability the true reward and transition probabilities
for all $s,a,k$. By~\Cref{lemma:optimism}, under these events, the optimistic
$Q$-values then satisfy the Bellman-type bounds
\begin{align*}
    &0 \leq \tilde{Q}^{k,(+)}_h(s,a)-\Big(r(s,a)+\sum_{s'\in\Scal}P(s'\mid s,a)\max_{a'\in\Acal}\tilde{Q}^{k,(+)}_{h+1}(s',a')\Big) \leq b_k(s,a,h)\\
    &\tilde{Q}^{k,{+}}_h(s,a) \leq H-h+1
\end{align*}
where $b_k(s,a,h)=\min(H-h+1,18S\beta_K(s,a)H/\max(1,N^k(s,a)-4S\beta_K(s,a)))$.
As shown by~\Cref{lemma:optimism_V}, optimism reduces regret to bounding $\sum_{k,h}b_{k-1}(S^k_h,A^k_h)$ ; the
counting argument in \Cref{app:upper_bound_rl} gives the stated rate.
\end{proof}

The upper bound shown in~\Cref{thm:upper_bound_rl} is clearly sub-optimal in the
low privacy regime. As $\epsilon \rightarrow \infty$, the upper bound becomes
$\tilde{O}(H^2 S^2 \sqrt{A K})$ which is much worse than the optimal non-private
regret rate in $\Theta(H \sqrt{S A K})$.  
Furthermore, the private term in the upper bound of~\cite{garcelonLocalDifferentialPrivacy2021} is given by
$H^3 S^2 A\sqrt{K}/ \epsilon$ which is better than the bound of~\Cref{thm:upper_bound_rl} as soon as $\epsilon \geq H \sqrt{A}$.
However, when $\epsilon \leq H \sqrt{A} - 2$, the bound of~\Cref{thm:upper_bound_rl} dominates and when $\epsilon \leq 1$, the gain is
of order $H \sqrt{A}$. The gain of the $\sqrt{A}$ factor is linked to the fact that we send a deterministic version of the policy and 
that we privatize a randomly chosen triplet from the trajectory. 
Indeed, privatizing a quantity such as $\sum_{h \in [H]} \ind{S^k_h=s, A^k_h=a, S^k_{h+1}=s'}$ by adding Laplace noise as done in~\cite{garcelonLocalDifferentialPrivacy2021}
gives an estimator of variance $H^2 / \epsilon^2$. Instead, privatizing $\ind{S^k_{J^k}=s, A^k_{J^k}=a, S^k_{J^k+1}=s'}$ can be done with a noise 
of scale $1/\epsilon^2$ and the variance of the resulting estimator is on the order of $H / \epsilon^2 \sum_{h=1}^H \ind{\pi^k_h(s)=a}$ which can be much smaller for 
actions rarely selected.  The gain of the $H$ factor is due to a tighter analysis. 
In~\cite{garcelonLocalDifferentialPrivacy2021}, the summation of the bonuses introduces an additional factor on 
top of the one due to the Bellman updates and the one due to the privacy-noise variance (see their Equation (50)). 
In contrast, our careful handling of the counts $\sum_{k,h}b_{k-1}(S^k_h,A^k_h)$ in the proof of~\Cref{thm:upper_bound_rl} 
allows us not to pay this additional factor.

\section{Conclusion and future work}
\label{sec:conclusion}
We improved regret bounds for tabular RL under local differential privacy. For
contextual bandits, the minimax regret is characterized up to constants and a
$\sqrt{\log(A)}$ factor~\Cref{thm:lower_bound,thm:upper_bound}. For finite-horizon RL, our reduction~\Cref{cor:lower_bound} gives a lower
bound that is $H$ times the contextual bandit lower bound. The improvement
over previous results increases as $\epsilon$ decreases, yielding a $\sqrt{S}$
improvement when $\epsilon \leq 1$. Our UCBVI-OUE upper bound~\Cref{thm:upper_bound_rl} improves over
previous results when $\epsilon \leq \max(H \sqrt{A} - 2, 1)$, which includes
the high privacy regime $\epsilon \leq 1$, where a $H\sqrt{A}$ improvement is obtained. 
We note that when $\epsilon \rightarrow \infty$ the upper bound of~\Cref{thm:upper_bound_rl} becomes
$\tilde{O}(H^2 S^2 \sqrt{A K})$ which is much worse than the optimal non-private
regret rate in $\Theta(H \sqrt{S A K})$. In fact, when $\epsilon \geq H \sqrt{A}$, 
the approach of~\cite{garcelonLocalDifferentialPrivacy2021} is preferable.

The main open question is whether the lower bound of~\Cref{cor:lower_bound} is
tight. In non-private problems, contextual bandits and tabular RL differ by a
factor $H$, and it is natural to conjecture that this remains true under LDP.
How to close the remaining $HS$ gap in the high-privacy regime is then the most
pressing 
question. We conjecture that closing the $S$-factor gap requires model-free
approaches, as model-based methods struggle with the high dimensionality of the transition kernel. However, existing analyses of model-free algorithms produce suboptimal rates even without privacy constraints. These challenging questions remain open for future investigation.

\clearpage
\bibliographystyle{abbrvnat}
\bibliography{biblio}


\appendix

\section{Proof of \Cref{thm:lower_bound}}
\label{app:lower_bound}

We prove \Cref{thm:lower_bound} for the following class of mechanisms, which
contains the mechanisms of~\Cref{sec:setting}.

\begin{definition}[Sequentially interactive $\epsilon$-LDP mechanism]
\label{def:sequentially_interactive}
Following~\cite{duchiLocalPrivacyStatistical2013}, a \emph{sequentially
interactive} mechanism is a sequence of channels $\mathcal{M} = (Q^k)_{k \in [K]}$
where, for each trajectory $\tau \in \mathbb{T}$ and each sequence of past
messages $y \in \Ycal^{k-1}$, $Q^k(\cdot \mid \tau, y)$ is a probability
distribution over $\Ycal$. At episode $k$, user $k$ sends
$M^k \sim Q^k(\cdot \mid \tau^k, (M^1, \ldots, M^{k-1}))$, drawn independently of
everything else given $\tau^k$ and $(M^1, \ldots, M^{k-1})$. The mechanism is
$\epsilon$-LDP if, for every $k \in [K]$ and every $y \in \Ycal^{k-1}$, the
channel $Q^k(\cdot \mid \cdot, y)$ is $\epsilon$-LDP: for any two trajectories
$\tau, \tau'$ and any measurable subset $Y \subset \Ycal$,
\begin{align*}
  Q^k(Y \mid \tau, y) \leq e^{\epsilon} Q^k(Y \mid \tau', y).
\end{align*}
\end{definition}

The (non-interactive) mechanisms of~\Cref{sec:setting} are the special case
where $Q^k(\cdot \mid \tau, y)$ is the distribution of $\mathcal{M}(\tau)$ for
every $k$ and $y$. Since the policy $\pi^k$ of a deterministic algorithm is a
function of $(M^1, \ldots, M^{k-1})$, the channel used at episode $k$ may in
particular depend on $\pi^k$. The regret is defined as in~\Cref{sec:setting},
with $M^k$ drawn as above.

\begin{proof}
  Fix $\Delta > 0$, we will choose its value later. Given $\phi \in
  \Acal^{\Scal}$ and $\sigma \in \Scal^K$, we define an instance $I_{\phi, \sigma} =
  (r_{\phi, \sigma}, \rho_{\phi, \sigma})$ by 
  \begin{align*}
    r_{\phi, \sigma}(s, a) = \frac12 + \Delta \ind{a=\phi(s)}, \quad
    \rho^k_{\phi, \sigma}(s) = \ind{s = \sigma_k}
  \end{align*} 

  Then consider the random instance $I = (r_I, \rho_I) = I_{\alpha, c}$ 
  where $\alpha(s)$ is chosen uniformly at random from $\Acal$ for each $s \in
  \Scal$ and $c_k$ is chosen uniformly at random from $\Scal$ for each $k \in
  [K]$. In other words, $I$ is uniformly distributed over the set of instances
  $\{I_{\phi, \sigma} : \phi \in \Acal^{\Scal}, \sigma \in \Scal^K\}$.
  Lastly, consider instance $I^0_s = (r^0_s, \rho^0_s)$ 
  defined as $I$ except that $r^0_{s}(s, a) = \frac12$ for each 
  $a \in \Acal$ and $r^0_{s}(s', a) = r_{I}(s', a)$ 
  for each $s' \neq s$ and $a \in \Acal$. 

Fix a deterministic algorithm $\mathfrak{A}$ and a sequentially interactive
$\epsilon$-LDP mechanism $\mathcal{M} = (Q^k)_{k \in [K]}$
(\Cref{def:sequentially_interactive}). We will show a lower bound on the expected regret of $(\mathfrak{A},
\mathcal{M})$ on the random instance $I$ and then conclude the proof by application of Yao's
minimax  principle~\cite{yaoProbabilisticComputationsUnified1977}.

  The expected regret can be expressed as
  \begin{align}
    &\EE[\Rcal(\mathfrak{A}, \Mcal, I)] \notag \\
    & = \EE \big[ \sum_{k=1}^K \left( V^*(S^k) - V^{\pi^k}(S^k) \right) \big] \notag \\
    &= \sum_{s \in \Scal} \EE \big[ \sum_{k=1}^K \ind{S^k = s} \left( V^*(s) - V^{\pi^k}(s) \right) \big] \notag \\
    &= \sum_{s \in \Scal} \EE \big[ \sum_{k=1}^K \ind{S^k = s} \Delta \ind{A^k \neq \alpha(s)} \big] \notag \\
    &= \Delta\bigg(K - \sum_{s \in \Scal} \EE\bigg[\sum_{k=1}^K \ind{S^k = s, A^k = \alpha(s)}\bigg]\bigg) \notag \\
    &= \Delta\bigg(K - \sum_{s \in \Scal} \sum_{j\in \Acal} \EE\bigg[\sum_{k=1}^K  \ind{A^k = j} \mid S^k=s, \alpha(s) = j\bigg] \frac1{SA} \bigg) \notag \\
    &= \Delta\bigg(K - \sum_{s \in \Scal} \sum_{j\in \Acal}  \EE\bigg[\underbrace{\sum_{k=1}^K \EE\bigg[ \ind{A^k = j} \mid S^k=s, (M^i)_{i \in [k-1]}\bigg]}_{N_{s, j}} \mid \alpha(s) = j\bigg] \frac1{SA} \bigg). \label{eq:regret_decomposition}
  \end{align}

  Call $\nu_{s, j}$ and $\nu_{s, 0}$ the distributions of the sequence of
  privatized messages $(M^i)_{i \in [K]}$ on instance $I$ conditioned on
  $\alpha(s) = j$ and on instance $I^0_s$  respectively, and denote $\EE_{s, j}$ and
  $\EE_{s, 0}$ the corresponding expectations.

  Since $N_{s, j}$ is bounded by $K$ and is a function of the sequence of privatized messages
  $(M^i)_{i \in [K]}$, from~\cite{auerNonstochasticMultiarmedBandit2002} (see also~\cite[Exercise
  15.2]{lattimoreBanditAlgorithms2020a}), we have
  \begin{align}
    \EE_{s, j}\bigg[N_{s, j}\bigg] \leq \EE_{s, 0}\bigg[N_{s, j}\bigg] + K \sqrt{\frac{1}{2} D_{\text{KL}}(\nu_{s, 0} \| \nu_{s, j})}.\label{eq:kl_inequality}
  \end{align}

Let us now detail the KL divergence $D_{\text{KL}}(\nu_{s, 0} \| \nu_{s,
j})$. To simplify notation, we write KL between random variables to mean the KL
between the distributions they are sampled from. Furthermore, for a random
variable $X$ that depends on the instance we consider, we write $ D_{\text{KL}}(X \mid
\alpha(s) = 0 \| X \mid \alpha(s) = j)$ to mean the KL divergence between the
distribution of $X$ when the instance is sampled from $I^0_s$ and the
distribution of $X$ when the instance is sampled from $I \mid \alpha(s) = j$.

For any $\ell \in [K]$, let $H^\ell = (M^i)_{i=1}^\ell$, we have
\begin{align}
    &D_{\text{KL}}((M^i)_{i \in [K]} \mid \alpha(s) = 0 \| (M^i)_{i \in [K]} \mid \alpha(s) = j) \notag \\
  &= \sum_{\ell = 0}^{K-1} \EE_{s, 0}\bigg[ D_{\text{KL}}(M^{\ell+1} \mid \alpha(s) = 0, H^{\ell} \| M^{\ell+1} \mid \alpha(s) = j, H^{\ell}) \bigg].\label{eq:kl_decomposition}
\end{align}
where the last equality is a direct consequence of the chain rule for KL
divergence. We now detail the KL divergence term $D_{\text{KL}}(M^{\ell+1} \mid
\alpha(s) = 0, H^{\ell} \| M^{\ell+1} \mid \alpha(s)  = j, H^{\ell})$.
Let $P^{\ell+1}_0$ and $P^{\ell+1}_j$ denote the distributions of
$(S^{\ell + 1}, A^{\ell + 1}, R^{\ell + 1}) \mid \alpha(s) = 0, H^\ell$ and of
$(S^{\ell + 1}, A^{\ell + 1}, R^{\ell + 1}) \mid \alpha(s) = j, H^\ell$.
Conditionally on $H^\ell$, the message $M^{\ell+1}$ is obtained by passing
$(S^{\ell + 1}, A^{\ell + 1}, R^{\ell + 1})$ through the channel
$Q^{\ell+1}(\cdot \mid \cdot, H^\ell)$. Hence, the distributions of
$M^{\ell+1} \mid \alpha(s) = 0, H^\ell$ and of $M^{\ell+1} \mid \alpha(s) = j, H^\ell$
are $P^{\ell+1}_0 Q^{\ell+1}$ and $P^{\ell+1}_j Q^{\ell+1}$, where $P Q^{\ell+1}$
denotes the distribution of the output of $Q^{\ell+1}(\cdot \mid \cdot, H^\ell)$
when its input is drawn from $P$. The same channel appears in both
distributions, and it is $\epsilon$-LDP by~\Cref{def:sequentially_interactive}.
Therefore, \cite[Theorem 2]{asoodehContractionLocallyDifferentially2024} gives
\begin{align}
    &D_{\text{KL}}(M^{\ell+1} \mid \alpha(s) = 0, H^{\ell} \| M^{\ell+1} \mid \alpha(s)  = j, H^{\ell}) \notag \\
  &\leq 4\frac{(\exp(\epsilon) - 1)^2}{\exp(\epsilon) }  D_{\text{TV}}(S^{\ell + 1}, A^{\ell + 1}, R^{\ell + 1} \mid \alpha(s) = 0, H^{\ell} \| S^{\ell + 1}, A^{\ell + 1}, R^{\ell + 1} \mid \alpha(s)  = j, H^{\ell})^2 \label{eq:tv_bound}
\end{align}

We now upper bound the TV divergence term. Notice that, the
distribution of $S^{\ell + 1}$ is independent of $\alpha$ and $H^\ell$, and the
distribution of $A^{\ell + 1}$ given $S^{\ell + 1}$ and $H^\ell$ is also
independent of $\alpha$ since $\pi^{\ell+1}$ is a function of $H^\ell$.
Furthermore, given $S^{\ell + 1}$ and $A^{\ell + 1}$, the reward $R^{\ell+1}$ is
independent of $H^\ell$. Since both instances share the optimal actions
$\phi(s')$ at the states $s' \neq s$, its distribution is the same under both
instances unless $S^{\ell + 1} = s$ and $A^{\ell + 1} = j$, in which case it is
$\Bcal(\frac12)$ under $I^0_s$ and $\Bcal(\frac12 + \Delta)$ under
$I \mid \alpha(s) = j$.
Therefore, we have
\begin{align}
  &D_{\text{TV}}(S^{\ell + 1}, A^{\ell + 1}, R^{\ell + 1} \mid \alpha(s) = 0, H^{\ell} \| S^{\ell + 1}, A^{\ell + 1}, R^{\ell + 1} \mid \alpha(s) = j, H^{\ell}) \notag \\
  &= \frac12 \sum_{x \in \Scal, a \in \Acal, r \in \{0, 1\}} \Big|\PP(S^{\ell + 1}=x, A^{\ell + 1}=a, R^{\ell + 1}=r \mid \alpha(s)=0, H^{\ell}) \notag \\
  & - \PP(S^{\ell + 1}=x, A^{\ell + 1}=a, R^{\ell + 1}=r \mid \alpha(s)=j, H^{\ell})\Big| \notag \\
  &= \frac12 \sum_{x \in \Scal, a \in \Acal, r \in \{0, 1\}} \PP(S^{\ell + 1} = x) \PP(A^{\ell + 1} = a \mid S^{\ell + 1}=x, H^\ell) \notag \\
  &\Big|\PP( R^{\ell + 1}=r \mid S^{\ell + 1}=x, A^{\ell + 1}=a,\alpha(s)=0, H^{\ell})  \notag \\
  &- \PP( R^{\ell + 1}=r \mid S^{\ell + 1}=x, A^{\ell + 1}=a,\alpha(s)=j, H^{\ell})\Big| \notag \\
  &= \PP(S^{\ell + 1} = s) \PP(A^{\ell + 1} = j \mid S^{\ell + 1}=s, H^\ell) D_{\text{TV}}(\Bcal(\frac12), \Bcal(\frac12 + \Delta)) \notag \\
  &= \frac{\Delta}{S} \PP(A^{\ell + 1} = j \mid S^{\ell + 1}=s, H^\ell) \label{eq:tv_computation}
  \end{align}

  Plugin the above upper bound into~\eqref{eq:tv_bound} we get
  \begin{align}
    &D_{\text{KL}}(M^{\ell+1} \mid \alpha(s)= 0, H^{\ell} \| M^{\ell+1} \mid \alpha(s)  = j, H^{\ell}) \notag \\
    &\leq \frac{4(\exp(\epsilon) - 1)^2}{\exp(\epsilon)}  \Delta^2 \frac1{S^2} \PP( A^{\ell + 1} = j \mid S^{\ell + 1} = s, H^\ell)^2 \label{eq:kl_bound}
  \end{align}

  Then, plugin the above upper bound into the chain rule decomposition of KL
  divergence \Cref{eq:kl_decomposition} we get 
  \begin{align*}
    &D_{\text{KL}}((M^i)_{i \in [K]} \mid \alpha(s) = 0 \| (M^i)_{i \in [K]} \mid \alpha(s) = j) \\
    &\leq 4\frac{(\exp(\epsilon) - 1)^2}{\exp(\epsilon)}  \Delta^2 \frac1{S^2} \sum_{\ell = 0}^{K-1} \EE_{s, 0}\bigg[ \PP( A^{\ell + 1} = j \mid S^{\ell + 1} = s, H^\ell)^2 \bigg].
  \end{align*}

  Lastly, plugin the above upper bound into the KL
  inequality~\eqref{eq:kl_inequality} we get
  \begin{align*}
    \EE_{s, j}\bigg[N_{s, j}\bigg] &\leq \EE_{s, 0}\bigg[N_{s, j}\bigg] \\
    &\quad + K \sqrt{2 \frac{(\exp(\epsilon) - 1)^2}{\exp(\epsilon)}  \Delta^2 \frac1{S^2} \sum_{\ell = 0}^{K-1} \EE_{s, 0}\bigg[ \PP( A^{\ell + 1} = j \mid S^{\ell + 1} = s, H^\ell)^2 \bigg]}.
  \end{align*}

  Summing over $j \in \Acal$, using Cauchy-Schwarz inequality and the fact that $\sum_{j \in \Acal} \PP(
  A^{\ell + 1} = j \mid S^{\ell + 1} = s, H^\ell)^2 \leq 1$, we get 
  \begin{align*}
    \sum_{j \in \Acal} \EE_{s, j}\bigg[N_{s, j}\bigg] &\leq \sum_{j \in \Acal} \EE_{s, 0}\bigg[N_{s, j}\bigg] + K  \sqrt{ 2 A  \sum_{\ell = 0}^{K-1}\frac{(\exp(\epsilon) - 1)^2}{\exp(\epsilon) } \Delta^2 \frac1{S^2}} \\
    &= K + K \sqrt{2 A K \frac{(\exp(\epsilon) - 1)^2}{\exp(\epsilon)} \Delta^2 \frac1{S^2}}.
  \end{align*}

  Then, we set $\Delta = \min(\frac12, \frac{\sqrt{A}}{4\sqrt{2 K \frac{(\exp(\epsilon) - 1)^2}{\exp(\epsilon)} \frac1{S^2}}})$, the above bound becomes
  \begin{align*}
    \sum_{j \in \Acal} \EE_{s, j}\bigg[N_{s, j}\bigg] &\leq K + \frac14 AK
  \end{align*}

  Finally, plugin this upper bound into the regret
  decomposition~\eqref{eq:regret_decomposition} we get
  \begin{align*}
    \EE[\Rcal(\mathfrak{A}, \Mcal, I)] &\geq \Delta\bigg(K - \sum_{s \in \Scal} \sum_{j\in \Acal}  \EE_{s, j}\bigg[N_{s, j} \bigg] \frac1{SA} \bigg) \\
    &\geq \Delta\bigg(K - S \frac{K + \frac14 AK}{SA} \bigg) \\
    &\geq \Delta\bigg(\frac34 K - K / A \bigg) \\
    &\geq \Delta\bigg(\frac34 K - K / 2 \bigg) \\
    &\geq \Delta\bigg(\frac14 K  \bigg) \\
    &\geq \min(\frac{K}{8}, \frac{  S \sqrt{A K \exp(\epsilon)}}{16\sqrt{2} (\exp(\epsilon) - 1)}) \\
    &\geq \frac1{32} \min(K, \frac{  S \sqrt{A K (\exp(\epsilon) + 1)}}{(\exp(\epsilon) - 1)})
  \end{align*}

  Then we re-start the proof again and develop the same argument but replacing
  the bound used in \Cref{eq:tv_bound} by the KL contraction
  of~\cite[Theorem 1]{asoodehContractionLocallyDifferentially2024}, applied to the
  same $\epsilon$-LDP channel $Q^{\ell+1}(\cdot \mid \cdot, H^\ell)$:
  \begin{align*}
    &D_{\text{KL}}(M^{\ell+1} \mid \alpha(s)= 0, H^{\ell} \| M^{\ell+1} \mid \alpha(s)  = j, H^{\ell})  \\
    &\leq \frac{(\exp(\epsilon) - 1)^2}{(\exp(\epsilon) + 1)^2} D_{\text{KL}}(S^{\ell + 1}, A^{\ell + 1}, R^{\ell + 1} \mid \alpha(s) = 0, H^{\ell} \| S^{\ell + 1}, A^{\ell + 1}, R^{\ell + 1} \mid \alpha(s) = j, H^{\ell}) \\
    &= \frac{(\exp(\epsilon) - 1)^2}{(\exp(\epsilon) + 1)^2 } \\
    &\EE[D_{\text{KL}}(R^{\ell + 1} \mid \alpha(s) = 0, H^{\ell}, S^{\ell + 1}, A^{\ell + 1} \| R^{\ell + 1} \mid \alpha(s) = j , H^{\ell}, S^{\ell + 1}, A^{\ell + 1}) \mid \alpha(s) = 0, H^{\ell} ] \\
&= \frac{(\exp(\epsilon) - 1)^2}{(\exp(\epsilon) + 1)^2} \PP(S^{\ell + 1} = s, A^{\ell + 1} = j \mid H^\ell, \alpha(s) = 0) D_{\text{KL}}(\Bcal(\frac12) \| \Bcal(\frac12 + \Delta)) \\
&\leq 4 \frac{(\exp(\epsilon) - 1)^2}{(\exp(\epsilon) + 1)^2}  \Delta^2 \frac1{S} \PP(A^{\ell + 1} = j \mid H^\ell, S^{\ell + 1}=s )
  \end{align*}
  where the last inequality is a consequence of the fact that
  $D_{\text{KL}}(\Bcal(\frac12) \| \Bcal(\frac12 + \Delta)) \leq 4 \Delta^2$ for
  $\Delta \leq \frac14$.
  The bound we obtain is analogous to the one given in~\Cref{eq:kl_bound}.

  Proceeding as before by plugin the above upper bound in the chain rule decomposition of KL
  divergence \Cref{eq:kl_decomposition}, then into the KL
  inequality~\Cref{eq:kl_inequality}, summing over $j$ using the identity $\sum_{j \in \Acal} \PP(A^{\ell + 1} = j \mid S^{\ell + 1} = s, H^\ell) = 1$ and applying
  Cauchy-Schwarz, we get  
  \begin{align*}
    \sum_{j \in \Acal} \EE_{s, j}\bigg[N_{s, j}\bigg] &\leq \sum_{j \in \Acal} \EE_{s, 0}\bigg[N_{s, j}\bigg] + K  \sqrt{ 2 A  \sum_{\ell = 0}^{K-1}\frac{(\exp(\epsilon) - 1)^2}{(\exp(\epsilon) + 1)^2} \Delta^2 \frac1{S}} \\
    &= K + K \sqrt{2 A K \frac{(\exp(\epsilon) - 1)^2}{(\exp(\epsilon) + 1)^2} \Delta^2 \frac1{S}}.
  \end{align*}
  Taking $\Delta = \min(\frac14, \frac{\sqrt{A}}{4\sqrt{2 K
  \frac{(\exp(\epsilon) - 1)^2}{(\exp(\epsilon) + 1)^2} \frac1{S}}})$, we get
  \begin{align*}
    \sum_{j \in \Acal} \EE_{s, j}\bigg[N_{s, j}\bigg] &\leq K + \frac14 AK
  \end{align*}
  Then the proof follows the same steps and yield the bound
  \begin{align*}
    \EE[\Rcal(\mathfrak{A}, \Mcal, I)] &\geq \min(\frac{K}{16}, \frac{   \sqrt{S A K } (\exp(\epsilon) + 1)}{16\sqrt{2} (\exp(\epsilon) - 1)}) \\
    &\geq \frac1{32}\min(K, \frac{   \sqrt{S A K } (\exp(\epsilon) + 1)}{(\exp(\epsilon) - 1)}) \\
  \end{align*}
  Putting the two bounds together we get
  \begin{align*}
    \EE[\Rcal(\mathfrak{A}, \Mcal, I)] &\geq \frac1{32} \min(K,  \frac{ \sqrt{\max( S, \exp(\epsilon) + 1) S A K (\exp(\epsilon) + 1)}}{\exp(\epsilon) - 1})
  \end{align*}

\end{proof}

\section{Proof of \Cref{thm:upper_bound}}
\label{app:upper_bound}

\begin{proof}
  The proof closely follows the standard analysis of Exp3~\cite[Theorem
  3.1]{auerNonstochasticMultiarmedBandit2002}.

    Form the potential 
    \[
    \Phi_k = \sum_{s \in \Scal} \ln\left(\sum_{a \in \Acal} w_k(s, a)\right).
    \]
    We have
    \begin{align*}
    \Phi_{k+1} - \Phi_k &= \sum_{s \in \Scal} \ln\left(\frac{\sum_{a \in \Acal} w_{k+1}(s, a)}{\sum_{a \in \Acal} w_k(s, a)}\right)\\
    &= \sum_{s \in \Scal} \ln\left(\sum_{a \in \Acal} \frac{w_k(s, a)}{\sum_{a' \in \Acal} w_k(s, a')} \exp(\eta \hat{V}^k(s, a))\right)\\
    \end{align*}
    
    Then notice that $\hat{V}^k(s, a) \in  \left[0, \frac{A }{\gamma} \right]$. We choose $\eta$
     such that $\gamma = A \eta \leq \frac12 $ so that $\eta \hat{V}^k(s, a) \in
     [0, 1]$. 
     We can then can apply successively the inequalities $\exp(x) \leq 1 + x + (e-2)x^2$ for $x \in [0, 1]$
     and $\ln(1 + x) \leq x$ for $x \in \RR$ to get
     \begin{align*}
      &\Phi_{k+1} - \Phi_k \\
      &\leq \sum_{s \in \Scal} \ln\bigg(1 + \sum_{a \in \Acal} \frac{w_k(s, a)}{\sum_{a' \in \Acal} w_k(s, a')}\eta \hat{V}^k(s, a) + \sum_{a \in \Acal} \frac{w_k(s, a)}{\sum_{a' \in \Acal} w_k(s, a')} (e - 2)\eta^2 \hat{V}^k(s, a)^2\bigg) \\
      &\leq \sum_{s \in \Scal} \sum_{a \in \Acal} \frac{w_k(s, a)}{\sum_{a' \in \Acal} w_k(s, a')}\eta \hat{V}^k(s, a) + \sum_{s \in \Scal}\sum_{a \in \Acal} \frac{w_k(s, a)}{\sum_{a' \in \Acal} w_k(s, a')} (e-2)\eta^2 \hat{V}^k(s, a)^2 \\
      &\leq \sum_{s \in \Scal} \sum_{a \in \Acal} \frac{\Pi^k(s, a)}{1 - \gamma}\eta \hat{V}^k(s, a) + \sum_{s \in \Scal}\sum_{a \in \Acal} \frac{\Pi^k(s, a)}{1 - \gamma} (e-2)\eta^2 \hat{V}^k(s, a)^2
      \end{align*}

      Then, summing over $k$, we get
      \begin{align*}
        &\Phi_{K+1} - \Phi_1 \\
        &\leq \sum_{k=1}^K \sum_{s \in \Scal} \sum_{a \in \Acal} \frac{\Pi^k(s, a)}{1 - \gamma}\eta \hat{V}^k(s, a) + \sum_{k=1}^K \sum_{s \in \Scal}\sum_{a \in \Acal} \frac{\Pi^k(s, a)}{1 - \gamma} (e-2)\eta^2 \hat{V}^k(s, a)^2  
      \end{align*}

      We have 
      $\Phi_1 = \sum_{s \in \Scal} \ln\left(\sum_{a \in \Acal} w_1(s, a)\right)
      = S \ln(A)$ and 
      \begin{align*}
       \Phi_{K+1} &= \sum_{s \in \Scal} \ln\left(\sum_{a \in \Acal} w_{K+1}(s, a)\right) \\
       &= \sum_{s \in \Scal} \ln\left(\sum_{a \in \Acal} \exp\left(\eta \sum_{k=1}^K \hat{V}^k(s, a)\right)\right)\\
       &\geq  \sum_{s \in \Scal} \ln\left(\exp\left(\eta \sum_{k=1}^K \hat{V}^k(s, a^*(s))\right)\right)\\
       &= \sum_{s \in \Scal} \eta \sum_{k=1}^K \hat{V}^k(s, a^*(s))
      \end{align*}
      where $a^*(s) = \argmax_{a \in \Acal} r(s, a)$.

      We therefore get the following identity
      \begin{align}
        &\sum_{s \in \Scal} \sum_{k=1}^K \hat{V}^k(s, a^*(s))  \notag \\
        &\leq \frac{S \ln(A)}{\eta} + \sum_{k=1}^K \sum_{s \in \Scal} \sum_{a \in \Acal} \frac{\Pi^k(s, a)}{1 - \gamma} \hat{V}^k(s, a) + \sum_{k=1}^K \sum_{s \in \Scal}\sum_{a \in \Acal} \frac{\Pi^k(s, a)}{1 - \gamma} (e-2)\eta \hat{V}^k(s, a)^2 \notag \\
        &= \sum_{k=1}^K \sum_{s \in \Scal} \sum_{a \in \Acal} \Pi^k(s, a) \hat{V}^k(s, a) \notag \\
        &+ \frac{S \ln(A)}{\eta} + \frac{\gamma}{1 - \gamma}\sum_{k=1}^K \sum_{s \in \Scal} \sum_{a \in \Acal} \Pi^k(s, a) \hat{V}^k(s, a) + \sum_{k=1}^K \sum_{s \in \Scal}\sum_{a \in \Acal} \frac{\Pi^k(s, a)}{1 - \gamma} (e-2)\eta \hat{V}^k(s, a)^2 \label{eq:potential_bound}
      \end{align}

      The term  $\sum_{a \in \Acal} \Pi^k(s, a) \hat{V}^k(s, a)^2$ can be upper
      bounded as 
      \begin{align*}
     &\sum_{a \in \Acal} \Pi^k(s, a) \hat{V}^k(s, a)^2 \\
      &= \sum_{a \in \Acal} \Pi^k(s, a) M^k(s) \frac{\ind{a = \pi^k(s)}}{\Pi^k(a, s)} \hat{V}^k(s, a) \\ 
      &\leq \sum_{a \in \Acal} \hat{V}^k(s, a)
      \end{align*}

      Then, we compute $\EE[\hat{V}^k(s, a)]$.
      \begin{align*}
        \EE[\hat{V}^k(s, a)] &= \EE[ M^k(s) \frac{\ind{a = \pi^k(s)}}{\Pi^k(a, s)}] \\
        &= \EE[ \EE[M^k(s) \mid V^k(s)] \frac{\ind{a = \pi^k(s)}}{\Pi^k(a, s)}] \\
      \end{align*}
      
      The term $\EE[M^k(s) \mid V^k(s)]$ is given by
      \begin{align*}
         \EE[M^k(s) \mid V^k(s)]
        &=  V^k(s) \frac12 + (1 - V^k(s)) \frac{1}{\exp(\epsilon) + 1} \\
        &=  V^k(s) (\frac12 - \frac1{\exp(\epsilon) + 1}) + \frac{1}{\exp(\epsilon) + 1} \\
        &=  \ind{S^k = s} B^k (\frac12 - \frac1{\exp(\epsilon) + 1}) + \frac{1}{\exp(\epsilon) + 1} \\
      \end{align*}
      which yield
      \begin{align*}
        &\EE[\hat{V}^k(s, a)] \\
        &= \EE[ \frac{\ind{S^k = s, A^k=a}}{\Pi^k(a, s)} B^k (\frac12 - \frac1{\exp(\epsilon) + 1})] + \frac{1}{\exp(\epsilon) + 1} \\
        &= \EE[ \ind{S^k = s} ]r(s, a) (\frac12 - \frac1{\exp(\epsilon) + 1}) + \frac{1}{\exp(\epsilon) + 1} \\
      \end{align*}

      Taking expectations in \Cref{eq:potential_bound}, we get
      \begin{align*}
        & \big(\frac12 - \frac1{\exp(\epsilon) + 1}\big)\EE[\sum_{k=1}^K r(S^k, a^*(S^k)) - \sum_{k=1}^Kr(S^k, A^k)] \\
        &\leq \frac{S \ln(A)}{\eta} + \frac{\gamma}{1 - \gamma}\sum_{k=1}^K \big(\frac12 - \frac1{\exp(\epsilon) + 1}+ \frac{S}{\exp(\epsilon) + 1}\big) \\
        &+ \sum_{k=1}^K \frac{A}{1 - \gamma} (e-2)\eta \big(\frac12 - \frac1{\exp(\epsilon) + 1} + \frac{S}{\exp(\epsilon) + 1}\big) \\
        &\leq \frac{S \ln(A)}{\eta} + 2(e-1) K A\eta \big(\frac12 - \frac1{\exp(\epsilon) + 1} + \frac{S}{\exp(\epsilon) + 1}\big)
      \end{align*}
      where we used $\gamma = A \eta$ and $\gamma \leq \frac12$.

      So that \begin{align*}
        &\EE[\sum_{k=1}^K r(S^k, a^*(S^k)) - \sum_{k=1}^Kr(S^k, A^k)] \\
        &\leq \frac{S \ln(A)}{\eta \big(\frac12 - \frac1{\exp(\epsilon) +
        1}\big)} + \frac{2(e-1) K A\eta \big(\frac12 - \frac1{\exp(\epsilon) +
        1} + \frac{S}{\exp(\epsilon) + 1}\big)}{\big(\frac12 -
        \frac1{\exp(\epsilon) + 1}\big)}
      \end{align*}

      Choosing $\eta = \min(\frac1{2A}, \sqrt{\frac{S \ln(A)}{2(e-1) K A \big(\frac12 -
      \frac1{\exp(\epsilon) + 1} + \frac{S}{\exp(\epsilon) + 1}\big)}})$ yields
      in case $\eta < \frac1{2A}$,
      \begin{align}
        &\EE[\sum_{k=1}^K r(S^k, a^*(S^k)) - \sum_{k=1}^Kr(S^k, A^k)] \notag \\
        &= 2 \frac{\sqrt{2(e-1) K AS \ln(A)\big(\frac12 - \frac1{\exp(\epsilon) + 1} + \frac{S}{\exp(\epsilon) + 1}\big)} }{\big(\frac12 - \frac1{\exp(\epsilon) + 1}\big)} \notag \\
        &= 2 \frac{\sqrt{2(e-1) K AS \ln(A)\big( 2 (\exp(\epsilon) + 1)^2  + 4(S - 1)(\exp(\epsilon) + 1)\big)} }{\exp(\epsilon) - 1}\notag \\
        &\leq 8 \frac{\exp(\epsilon) + 1}{\exp(\epsilon) - 1}\sqrt{SK A  \ln(A) } +  8 \frac{\sqrt{\exp(\epsilon) + 1}}{\exp(\epsilon) - 1} S \sqrt{K A \ln(A)} \label{eq:regret_bound}
      \end{align}
      where we used 
      $\frac12 - \frac1{\exp(\epsilon) + 1} = \frac{\exp(\epsilon) - 1}{2(\exp(\epsilon) +
      1)}$. 

      If $\eta = \frac1{2A}$, then we have
\begin{align*}
  &\frac1{2A} \leq \sqrt{\frac{S \ln(A)}{2(e-1) K A \big(\frac12 - \frac1{\exp(\epsilon) + 1} + \frac{S}{\exp(\epsilon) + 1}\big)}} \\
  &\implies K \leq \frac{2 A S \ln(A)}{(e-1) \big(\frac12 - \frac1{\exp(\epsilon) + 1} + \frac{S}{\exp(\epsilon) + 1}\big)} \\
  &\implies K \leq \frac{4  A S \ln(A) (\exp(\epsilon) + 1)}{(e-1) \big(\exp(\epsilon) + 2 (S - 1)\big)} \\
  &\implies K \leq \frac{4  A S \ln(A) (\exp(\epsilon) + 1)}{ \big(\exp(\epsilon)- 1\big)} \\
\end{align*}
  Finally, notice that the regret is always upper bounded by $K$ and we can then
  write
  \begin{align*}
  K &= \sqrt{K} \sqrt{K} \\ 
  &\leq \sqrt{K} \sqrt{\frac{4  A S \ln(A) (\exp(\epsilon) + 1)}{ \big(\exp(\epsilon)- 1\big)}} \\
  &\leq 8 \frac{\exp(\epsilon) + 1}{\exp(\epsilon) - 1}\sqrt{SK A  \ln(A) }
  \end{align*}
therefore, the bound~\Cref{eq:regret_bound} holds in all cases.  

Finally, using the identity $x + y \leq 2 \max(x, y)$, we get
\begin{align*}
  &\EE[\sum_{k=1}^K r(S^k, a^*(S^k)) - \sum_{k=1}^Kr(S^k, A^k)] \\
  &\leq 16 \max\bigg( \frac{\exp(\epsilon) + 1}{\exp(\epsilon) - 1}\sqrt{SK A  \ln(A) }, \frac{\sqrt{\exp(\epsilon) + 1}}{\exp(\epsilon) - 1} S \sqrt{K A \ln(A)}\bigg) \\
  &\leq 16 \min(K,  \frac{ \sqrt{\max( S, \exp(\epsilon) + 1) S A K  (\exp(\epsilon) + 1) \ln(A)}}{\exp(\epsilon) - 1})
\end{align*}
\end{proof}

\section{Proof of \Cref{cor:lower_bound}}
\label{app:cor_lower_bound}
\begin{proof}
Fix an algorithm $\mathfrak{A}$ and an $\epsilon$-LDP mechanism $\mathcal{M}$
and consider the instance $\tilde{I} = (\tilde{\Scal}, \tilde{\Acal}, \tilde r,
\{ \tilde \rho^k\}_{k=1}^K)$ 
of \Cref{thm:lower_bound} that achieves high regret
in the private contextual bandit setting with $S-2$ states and $A$ arms. 
In instance $\tilde{I}$, rewards are sampled from a Bernoulli distribution with
mean $\tilde{r}(s, a)$ for each $s \in \tilde{\Scal}$ and $a \in \tilde{\Acal}$

We construct an instance $I'$ with $S$ states and $A$ actions
 of finite-horizon MDP 
$(\Scal, \Acal, P, r, \{\rho^k\}_{k=1}^K)$ as follows.  
The action space is $\Acal = \tilde{\Acal}$. 
The state space is $\Scal = \tilde{\Scal} \cup \{\text{good}, \text{bad}\}$ 
having two additional states. 
All states have reward zero, except the good state which has
reward one for every action. The good state and the bad state are absorbing.
The initial distributions are defined as 
$\rho^k(\text{good}) = \rho^k(\text{bad}) = 0$ and 
$\rho^k(s) = \tilde{\rho}^k(s)$ for each $s \in \tilde{\Scal}$.
Then, we choose the transition probabilities as 
\[
P(\text{good} \mid s, a) = \tilde r(s, a), \quad P(\text{bad} \mid s, a) = 1 - \tilde r(s, a)
\]
for each $s \in \tilde{\Scal}$ and $a \in \Acal$. 
Fix $s \in \Scal \setminus \{\text{good}, \text{bad}\}$, $a \in \Acal$, $k \in [K]$, fix
any policy $\pi$ and sample a trajectory $(S^k_h, A^k_h)_{h=1}^H$ such that $S^k_1 =
s$ and $A^k_1 = a$.
We have
\begin{align*}
\sum_{h \in [H]} r'(S^k_h, A^k_h) &= r'(s, a) + \sum_{h=2}^H r'(S^k_h, A^k_h) \\
&= r'(s, a) + (H - 1) \ind{S^k_2 = \text{good}} \\
&= (H - 1) \ind{S^k_2 = \text{good}}.
\end{align*}
Notice that $\ind{S^k_2 = \text{good}}$ is a Bernoulli random variable with mean
$\tilde{r}(s, a)$. Therefore, instance $I$ is equivalent to that of $\tilde{I}$ but with rewards
scaled by $H - 1$. Furthermore, the bandit mechanism induced by $\mathcal{M}$ is
$\epsilon$-LDP private since each bandit observation is mapped to a trajectory
and then privatized.
Therefore, the lower bound of \Cref{thm:lower_bound}
applies with a multiplicative factor of $H-1$ and we have 
    \begin{align*}
        \Rcal(\mathfrak{A}, \mathcal{M}, I) \geq \frac{H-1}{32} \min\bigg(K,  \frac{ \sqrt{\max( S-2, \exp(\epsilon) + 1) (S-2) A K (\exp(\epsilon) + 1)}}{\exp(\epsilon) - 1}\bigg)
    \end{align*}

Then since $H \geq 2$ and $S\geq 4$, we have $H-1 \geq \frac{H}{2}$ and $S-2
\geq \frac{S}{2}$ which concludes the proof.
\end{proof}

\section{Proof of \Cref{thm:upper_bound_rl}}
\label{app:upper_bound_rl}

\begin{proof}
  Define for each $s, a \in \Scal \times \Acal$, $b \in \{0, 1\}$ and $s' \in \Scal$,
  \[
  N^k(s, a, b, s') = \sum_{i \in [k]} \sum_{h \in [H]} \ind{S^i_h = s, A^i_h =
  a, B^i_h = b, S^i_{h+1} = s'}
  \]
  The following \Cref{lemma:deviation_hatN}
  quantifies the deviation of $\hat{N}^k(s, a, b, s')$ from $N^k(s, a, b, s' )$.
  \begin{lemma}
\label{lemma:deviation_hatN}
Let $\mathcal{E}_1$ be the event that for each $s \in \Scal, a \in \Acal, b \in
\{0, 1\}, s' \in \Scal, k \in [K]$ and $\delta \leq \frac1{e}$, the following holds
\begin{align*}
    &|\hat{N}^k(s, a, b, s') - N^k(s, a, b, s')| \\
    &\leq \sqrt{20 H \Upsilon_\epsilon^2 \sum_{h \in [H], i \in [k]} \ind{a = \pi^i_h(s)} \log(\frac{8K S^2 A\log(16K)}{\delta})} + 4 H \Upsilon_\epsilon \log(\frac{8K S^2 A \log(16K)}{\delta}) \\
    &\leq \underbrace{\sqrt{20 H \Upsilon_\epsilon^2 \sum_{h \in [H], i \in [k]} \ind{a = \pi^i_h(s)} \log(\frac{48K^2 S^2 A}{\delta})} + 4 H \Upsilon_\epsilon \log(\frac{48K^2 S^2 A }{\delta})}_{:=\beta_k(s, a)}
\end{align*}
Then $\PP(\mathcal{E}_1) \geq 1 - \delta$.
\end{lemma}

The proof of \Cref{lemma:deviation_hatN} relies on Freedman's inequality 
for martingales. We use the following version
from~\cite{hsuMixingTimeEstimation2015}[Corollary 1, with $c=2$, $x=\log(\frac{1}{\delta})$].
 \begin{theorem}[Freedman's inequality~\cite{hsuMixingTimeEstimation2015}]
  \label{theorem:freedman}
    Let $(X_i)_{i \in [K]}$ adapted to a filtration $(\Fcal_i)_{i \in [K]}$ 
    such that $\EE[X_i \mid \Fcal_{i-1}] = 0$ and $|X_i| \leq b$ for each $i \in
    [K]$ and $b \in \RR_+$.

    Define $S_k = \sum_{i=1}^k X_i$ and  
    $V_k = \sum_{i=1}^k \EE[X_i^2 \mid \Fcal_{i-1}]$ for each $k \in [K]$.
    Then,
    \begin{align*}
        \PP\left(\exists k \in [K]: S_k \geq \sqrt{ 4 V_k \log(\frac1{\delta})} + \frac{4}{3} b \log(\frac1{\delta}) \right) \leq (1 + \max(1, \ceil{\log(2K / \log(\frac1{\delta}))}))\delta
    \end{align*}
\end{theorem} 
from which we derive the following corollary.
\begin{corollary}
\label{corollary:freedman}
Under the same assumptions as \Cref{theorem:freedman}, if $\frac1{\delta} \geq
e$, we have,
\begin{align*}
    \PP\left(\exists k \in [K]: S_k \geq \sqrt{ 4 V_k \log(\frac{\log(16K)}{\delta})} + \frac{4}{3} b \log(\frac{\log(16K)}{\delta}) \right) \leq \delta
\end{align*}
\end{corollary}

\begin{proof}[Proof of \Cref{lemma:deviation_hatN}]
Define the martingale difference sequence $(X_k)_{k \in [K]}$ where 
\[
X_k = H \hat{I}^k(s, b, s') \ind{\pi^k_{J^k}(s) = a} - \sum_{h \in [H]} \ind{S^k_h = s, A^k_h =
  a, B^k_h = b, S^k_{h+1} = s'}.
\]
Define a filtration $(\Fcal_k)_{k \in [K]}$ adapted to $(X_k)_{k \in [K]}$ by
\[
\Fcal_k = \sigma((J^i, S^i_j, A^i_j, B^i_j, S^i_{j+1}, M^i)_{j \in [H], i \in [k]}).
\]
It holds that $\EE[X_k \mid \Fcal_{k-1}] = 0$ for each $k \in [K]$ and since
$|\hat{I}^k| \leq 2 \Upsilon_\epsilon$ for each $k \in [K]$, we have that
$|X_k| \leq 3 H \Upsilon_\epsilon$ for each $k \in [K]$. 
Moreover, we have
\begin{align*}
    &\EE[X_k^2 \mid \Fcal_{k-1}] \\
    &=  \EE\bigg[\bigg(H \hat{I}^k(s, b, s') \ind{\pi^k_{J^k}(s) = a} - \sum_{h \in [H]} 
    \ind{S^k_h = s, A^k_h = a, B^k_h = b, S^k_{h+1} = s'} \bigg)^2 \mid \Fcal_{k-1}\bigg] \\
  &\leq 5 H \Upsilon_\epsilon^2 \sum_{h \in [H]} \ind{a = \pi^k_h(s)}
\end{align*}

Then Freedman's inequality applied to $(X_k)_k$ and $(-X_k)_k$ along with a
union bound over $s, a, b, s' \in \Scal
\times \Acal \times \{0, 1\} \times \Scal \times [K]$ and over the two resulting
inequalities gives the proof.
\end{proof}



Then define $Q^*_h$ for each $(s, a , h) \in \Scal \times \Acal \times [H]$ by $Q^*_{H+1}(s, a) = 0$ and
\[
Q^*_h(s, a) = r(s, a) + \sum_{s' \in \Scal} P(s' \mid s, a) \max_{a' \in \Acal} Q^*_{h+1}(s', a') 
\]
Then, we show in \Cref{lemma:optimism} the following optimism property for
$\tilde{Q}^{k, (+)}$.
\begin{lemma}
\label{lemma:optimism}
With probability at least $1 - 3\delta$, for each $k \in [K]$, $s \in \Scal, h
\in [H]$ and $a \in \Acal$, we have
\begin{align}
    &0 \leq \tilde{Q}^{k, (+)}_h(s, a) - \big(r(s, a) + \sum_{s' \in \Scal} P(s' \mid s, a) \max_{a' \in \Acal} \tilde{Q}^{k, (+)}_{h+1}(s', a')\big) \leq b_k(s, a, h) \label{eq:opt1} \\
    &\tilde{Q}^{k, {+}}_h(s, a) \leq H - h + 1 \label{eq:opt2}
\end{align}
where 
\[
b_k(s, a, h) = \min(H - h + 1, \frac{18 S \beta_K(s, a) H}{\max(1, N^k(s, a) - 4 S \beta_K(s, a))}).
\]
\end{lemma}
\begin{proof}[Proof of \Cref{lemma:optimism}]
\Cref{eq:opt2} follows easily by induction. The case $H+1$ is satisfied since
$\tilde{Q}^{k, (+)}_{H+1}(s, a) = 0$ then assuming $\tilde{Q}^{k, (+)}_{h+1}(s, a) \leq H -
h$, we have since $\rho \leq 1$ and $\sum_{s' \in \Scal} \nu(s') = 1$ that
$\tilde{Q}^{k, (+)}_h(s, a) \leq H - h + 1$.

For \Cref{eq:opt1}, we first need to show the following \Cref{lemma:properties_confidence_set}
stating that $P$ and $r$ belong to the
confidence sets.
\begin{lemma}
\label{lemma:properties_confidence_set}
Call $\Ecal_2$ the event that for each $s \in \Scal$, $a \in \Acal, k \in [K]$, the following holds
\begin{align*}
  P(s, a, \cdot), r(s, a) \in \Ccal^k(s, a)
\end{align*}
Then $\PP(\Ecal_2 \cap \Ecal_1) \geq 1 - 3 \delta$.
\end{lemma}
\begin{proof}[Proof of \Cref{lemma:properties_confidence_set}]
  First if $\tilde{N}^{k, (-)}(s, a) = 1$ and $\Ecal_1$ holds, we have
  for each $s, a, k \in \Scal \times \Acal \times [K]$ that
  \begin{align*}
   \sum_{s' \in \Scal, b \in \{0, 1\}} N^k(s, a, b, s') &\leq \sum_{s' \in \Scal, b \in \{0, 1 \}} (\hat{N}^k(s, a, b, s') + \beta_k(s, a)) && \text{(By $\Ecal_1$)} \\
   &\leq  \sum_{s' \in \Scal, b \in \{0, 1 \}} (\hat{N}^k(s, a, b, s') + \beta_k(s, a)) \\
   &= \sum_{s' \in \Scal, b \in \{0, 1\}} \hat{N}^k(s, a, b, s') + 2 S \beta_k(s, a) \\
   &\leq \tilde{N}^{k, (-)}(s, a) + 4 S \beta_k(s, a) \\
   &= 1 + 4 S \beta_k(s, a)
  \end{align*}
and thus
\begin{align*}
   &|\tilde{N}^{k, (-)}(s, a) r(s, a) - \sum_{s' \in \Scal} \hat{N}^k(s, a, 1, s') | \\
   &= |r(s, a) - \sum_{s' \in \Scal} \hat{N}^k(s, a, 1, s') | \\
   &\leq 1 + |\sum_{s' \in \Scal} \hat{N}^k(s, a, 1, s')| \\
   &\leq 1 + \sum_{s' \in \Scal}| \hat{N}^k(s, a, 1, s')| \\
   &\leq 1 + \sum_{s' \in \Scal} N^k(s, a, 1, s') + \beta_k(s, a)) && \text{(By $\Ecal_1$ and $\beta_k(s, a) > 0$)} \\
   &\leq 2 + 5 S\beta_k(s, a) \\
   &\leq 7 S \beta_k(s, a)
\end{align*}
and
\begin{align*}
  & \|\tilde{N}^{k, (-)}(s, a) P(s, a, \cdot) - \sum_{b \in \{0, 1 \}} \hat{N}^k(s, a, b, \cdot) \|_1 \\
  &= \| P(s, a, \cdot) - \sum_{b \in \{0, 1 \}} \hat{N}^k(s, a, b, \cdot) \|_1 \\
  &\leq 1 + \sum_{b \in \{0, 1\}, s' \in \Scal} |\hat{N}^k(s, a, b, s')| \\
  &\leq 1 + \sum_{b \in \{0, 1\}, s' \in \Scal} (N^k(s, a, b, s') + \beta_k(s, a)) && \text{(By $\Ecal_1$)} \\
  &\leq 2 + 6 S \beta_k(s, a) \\
  &\leq 8 S \beta_k(s, a)
\end{align*}
which implies $\Ecal_2$.

We now study the case $\tilde{N}^{k, (-)}(s, a) > 1$.
Define the martingale difference sequence $(X_k)_{k \in [K]}$ where
  \[
  X_k = H \hat{I}^k(s, b, s') \ind{\pi^k_{J^k}(s) = a} - \sum_{h \in [H]} \ind{S^k_h = s, A^k_h =
  a, B^k_h = b} P(s' \mid s, a).
  \]
  Then, denoting 
  \[
  N^k(s, a, b) =
  \sum_{i \in [k]} \sum_{h \in [H]} \ind{S^i_h = s, A^i_h =
  a, B^i_h = b}
  \] and applying the same steps as in the proof of 
  \Cref{lemma:deviation_hatN}, we
  show that with probability at least $1 - \delta$, for each $s \in \Scal, a \in
  \Acal, b \in \{0, 1\}, s' \in \Scal, k \in [K]$, it holds that
  \[
  | N^k(s, a, b ) P(s, a, s') - \hat{N}^k(s, a, b, s')| \leq \beta_k(s, a).
  \]
  Equivalently, if we denote $\Ecal_3$ the event that for each $s, a, b, s', k \in \Scal \times \Acal \times \{0, 1\} \times \Scal \times [K]$, 
  \[
  | N^k(s, a, b ) P(s, a, s') - \hat{N}^k(s, a, b, s')| \leq \beta_k(s, a),
  \]
  then $\PP(\Ecal_3) \geq 1 - \delta$. 

  Consider $\Ecal_1$ and $\Ecal_3$ holds and $\tilde{N}^{k, (-)}(s, a) > 1$
  meaning 
  \[
  \tilde{N}^{k, (-)}(s, a) = \sum_{b \in \{0, 1\}, s' \in \Scal}  (\hat{N}^k(s, a, b, s') - \beta_k(s, a)).
  \]
  Then, we have that for each $s, a, s',k \in \Scal \times
  \Acal \times \Scal \times [K]$,
  \begin{align*}
    &\sum_{b \in \{0, 1\}}\hat{N}^k(s, a, b, s') \\
    &\geq \sum_{b \in \{0, 1\}} \big(N^k(s, a, b) P(s, a, s') - \beta_k(s, a) \big) && \text{(By $\Ecal_3$)}\\
    &\geq \sum_{b \in \{0, 1\}} \bigg(\sum_{y \in \Scal}\big(\hat{N}^k(s, a, b, y) - \beta_k(s, a)\big) P(s, a, s') - \beta_k(s, a) \bigg) && \text{(By $\Ecal_1$)} \\
    &= \tilde{N}^{k, (-)}(s, a) P(s, a, s') - 2 \beta_k(s, a) && \text{(By $\tilde{N}^{k, (-)}(s, a) > 1$)}
  \end{align*}

  Similarly, we can write for each $s, a, s', k \in \Scal \times \Acal \times
  \Scal \times [K]$,
  \begin{align*}
    &\sum_{b \in \{0, 1\}} \hat{N}^k(s, a, b, s') \\
    &\leq \sum_{b \in \{0, 1\}} \big(N^k(s, a, b) P(s, a, s') + \beta_k(s, a) \big) && \text{(By $\Ecal_3$)}\\
    &\leq \sum_{b \in \{0, 1\}} \bigg(\sum_{y \in \Scal}\big(\hat{N}^k(s, a, b, y) + \beta_k(s, a)\big) P(s, a, s') + \beta_k(s, a) \bigg) && \text{(By $\Ecal_1$)}\\
    &= \big(\tilde{N}^{k, (-)}(s, a) + 2 \sum_{b \in \{0, 1 \}, y \in \Scal} \beta_k(s, a) \big) P(s, a, s') + 2 \beta_k(s, a) && \text{(By $\tilde{N}^{k, (-)}(s, a) > 1$)} \\ 
    &\leq \tilde{N}^{k, (-)}(s, a) P(s, a, s') + 4 S \beta_k(s, a) P(s, a, s') + 2 \beta_k(s, a) \\
  \end{align*}
  so we get 
  \[
  |\tilde{N}^{k, (-)}(s, a) P(s, a, s') - \sum_{b \in \{0, 1\}} \hat{N}^k(s, a, b, s')| 
  \leq 4 S \beta_k(s, a) P(s, a, s') + 2 \beta_k(s, a)
  \]
  which implies
  \[
  \|\tilde{N}^{k, (-)}(s, a) P(s, a, \cdot) - \sum_{b \in \{0, 1\}} \hat{N}^k(s, a, b, \cdot)\|_1 \leq 6 S \beta_k(s, a)
  \]
  meaning $P(s, a, \cdot)$ satisfies the corresponding constraint of the confidence set $\Ccal^k(s, a)$.

  For the other constraint, we consider the martingale difference sequence
  $(X_k)_{k \in [K]}$ where
  \[
  X_k = H \hat{I}^k(s, 1, s') \ind{\pi^k_{J^k}(s) = a} - \sum_{h \in [H]} \ind{S^k_h = s, A^k_h =
  a, S^k_{h+1} = s'} r(s, a).
  \]
  and apply the same steps as in the proof of \Cref{lemma:deviation_hatN} to
  show that with probability at least $1 - \delta$, for each $s \in \Scal, a \in
  \Acal,k \in [K], s' \in \Scal$, it holds that 
  \[
  | N^k(s, a, s') r(s, a) - \hat{N}^k(s, a, 1, s') | \leq \beta_k(s, a).
  \]
  Equivalently, if we denote $\Ecal_4$ the event that for each $s, a, s', k \in
  \Scal \times \Acal \times \Scal \times [K]$,
  \[
  | N^k(s, a, s') r(s, a) - \hat{N}^k(s, a, 1, s') | \leq \beta_k(s, a),
  \]
  then $\PP(\Ecal_4) \geq 1 - \delta$.
   Consider $\Ecal_1$ and $\Ecal_4$ hold and $\tilde{N}^{k, (-)}(s, a) > 1$. We
   have that for each $s, a, s', k \in \Scal \times \Acal \times \Scal \times [K]$,
  \begin{align*}
    &\sum_{s' \in \Scal} \hat{N}^k(s, a, 1, s') \\
    &\geq \sum_{s' \in \Scal} \big(N^k(s, a, s') r(s, a) - \beta_k(s, a)\big) && \text{(By $\Ecal_4$)} \\
    &\geq  \sum_{s' \in \Scal} \bigg(\big(\sum_{b \in \{0, 1\}} \hat{N}^k(s, a, b, s') - 2\beta_k(s, a)\big) r(s, a) - \beta_k(s, a) \bigg) && \text{(By $\Ecal_1$)}\\
    &\geq \tilde{N}^{k, (-)}(s, a) r(s, a) - S \beta_k(s, a) && \text{(By $\tilde{N}^{k, (-)}(s, a) > 1$)}
  \end{align*}
  and 
  \begin{align*}
    &\sum_{s' \in \Scal} \hat{N}^k(s, a, 1, s') \\
    &\leq \sum_{s' \in \Scal} \big(N^k(s, a, s') r(s, a) + \beta_k(s, a)\big) && \text{(By $\Ecal_4$)}\\
    &\leq  \sum_{s' \in \Scal} \bigg(\sum_{b \in \{0, 1\}} \big(\hat{N}^k(s, a, b, s') + \beta_k(s, a)\big) r(s, a) + \beta_k(s, a) \bigg) && \text{(By $\Ecal_1$)}\\
    &\leq (\tilde{N}^{k, (-)}(s, a) + 4 S \beta_k(s, a)) r(s, a) + S\beta_k(s, a) && \text{(By $\tilde{N}^{k, (-)}(s, a) > 1$)}\\
    &\leq \tilde{N}^{k, (-)}(s, a) r(s, a) + 5 S \beta_k(s, a)
  \end{align*} 

  So it holds that $\Ecal_2 \supseteq \Ecal_1 \cap \Ecal_3 \cap \Ecal_4$ and
  therefore $\PP(\Ecal_2 \cap \Ecal_1) \geq 1 - 3 \delta$.
\end{proof}

In the rest of the proof, we assume $\Ecal_2 \cap \Ecal_1$ holds.
First, we show the left-hand side of \Cref{eq:opt1}. 
We have that for each $s, a \in \Scal \times \Acal$ and $h \in [H]$,
\begin{align*}
  &\tilde{Q}^{k, (+)}_h(s, a) - \big(r(s, a) + \sum_{s' \in \Scal} P(s' \mid s, a) \max_{a' \in \Acal} \tilde{Q}^{k, (+)}_{h+1}(s', a')\big) \\
  &= \max_{\nu, \rho \in \Ccal^k(s, a)} \bigg(\rho + \sum_{s' \in \Scal} \nu(s') \max_{a' \in \Acal} \tilde{Q}^{k, (+)}_{h+1}(s', a') \bigg) \\
  &- \bigg(r(s, a) + \sum_{s' \in \Scal} P(s' \mid s, a) \max_{a' \in \Acal} \tilde{Q}^{k, (+)}_{h+1}(s', a') \bigg) \\
  &\geq 0 && \text{(By $\Ecal_2$)}
\end{align*}

Then, we show the right-hand side of \Cref{eq:opt1}.
By $\Ecal_2$ and the definition of $\Ccal^k(s, a)$, it holds that for any $\nu,
\rho \in \Ccal^k(s, a)$,
\begin{align*}
&|\tilde{N}^{k, (-)}(s, a) (\rho - r(s, a)) | \\
&\leq |\tilde{N}^{k, (-)}(s, a) \rho - \sum_{s' \in \Scal} \hat{N}^k(s, a, 1, s') | + |\sum_{s' \in \Scal} \hat{N}^k(s, a, 1, s') - \tilde{N}^{k, (-)}(s, a) r(s, a)  | \\
&\leq 14 S \beta_K(s, a)
\end{align*}
Due to $\Ecal_1$, $\tilde{N}^{k, (-)}(s, a) \geq \max(1, N^k(s, a) - 4 S \beta_K(s, a))$ so
that we have
\[
|\rho - r(s, a) | \leq \frac{14 S \beta_K(s, a)}{\max(1, N^k(s, a) - 4 S \beta_K(s, a))} 
\]
Similarly for $\nu$, we have
\begin{align*}
&\|\tilde{N}^{k, (-)}(s, a) (\nu - P(s, a, \cdot))\|_1 \\
&\leq \|\tilde{N}^{k, (-)}(s, a) \nu - \sum_{b \in \{0, 1\}} \hat{N}^k(s, a, b, \cdot) \|_1 + \|\sum_{b \in \{0, 1\}} \hat{N}^k(s, a, b, \cdot) - \tilde{N}^{k, (-)}(s, a) P(s, a, \cdot) \|_1 \\
&\leq 16 S \beta_K(s, a)
\end{align*}
and due to $\Ecal_1$, $\tilde{N}^{k, (-)}(s, a) \geq \max(1, N^k(s, a) - 4 S
\beta_K(s, a))$ so that we have
\[
\|\nu - P(s, a, \cdot)\|_1 \leq \frac{16 S \beta_K(s, a)}{\max(1, N^k(s, a) - 4 S \beta_K(s, a))}
\]

Then, we conclude the proof by writing for each $s, a \in \Scal \times \Acal$ and $h \in [H]$,
\begin{align*}
&\tilde{Q}^{k, (+)}_h(s, a) - \big(r(s, a) + \sum_{s' \in \Scal} P(s' \mid s, a) \max_{a' \in \Acal} \tilde{Q}^{k, (+)}_{h+1}(s', a')\big) \\
&= \max_{\nu, \rho \in \Ccal^k(s, a)} \bigg(\rho + \sum_{s' \in \Scal} \nu(s') \max_{a' \in \Acal} \tilde{Q}^{k, (+)}_{h+1}(s', a') \bigg) \\
&- \bigg(r(s, a) + \sum_{s' \in \Scal} P(s' \mid s, a) \max_{a' \in \Acal} \tilde{Q}^{k, (+)}_{h+1}(s', a') \bigg) \\ 
&\leq \max_{\nu, \rho \in \Ccal^k(s, a)} \bigg(r(s, a) + \frac{14 S \beta_K(s, a)}{\max(1, N^k(s, a) - 4 S \beta_K(s, a))}  \\
&+ \sum_{s' \in \Scal} \nu(s') \max_{a' \in \Acal} \tilde{Q}^{k, (+)}_{h+1}(s', a') \bigg) \\
&- \bigg(r(s, a) + \sum_{s' \in \Scal} P(s' \mid s, a) \max_{a' \in \Acal} \tilde{Q}^{k, (+)}_{h+1}(s', a') \bigg) \\ 
&=  \frac{14 S \beta_K(s, a)}{\max(1, N^k(s, a) - 4 S \beta_K(s, a))} + \max_{\nu \in \Ccal^k(s, a)} \sum_{s' \in \Scal} (\nu(s') - P(s' \mid s, a)) \max_{a' \in \Acal} \tilde{Q}^{k, (+)}_{h+1}(s', a') \\
&\leq \frac{14 S \beta_K(s, a)}{\max(1, N^k(s, a) - 4 S \beta_K(s, a))} + \max_{\nu \in \Ccal^k(s, a)} \|\nu(\cdot) - P(s, a, \cdot)\|_1 (H - h) \\ 
&\leq \frac{14 S \beta_K(s, a)}{\max(1, N^k(s, a) - 4 S \beta_K(s, a))} + \frac{16 S \beta_K(s, a) (H - h)}{\max(1, N^k(s, a) - 4 S \beta_K(s, a))} \\
&\leq \frac{18 S \beta_K(s, a) H}{\max(1, N^k(s, a) - 4 S \beta_K(s, a))} \\
\end{align*}
Furthermore, since $\tilde{Q}^{k, (+)}_h(s, a) \leq H - h + 1$ and 
\[
\tilde{Q}^{k, (+)}_h(s, a)-\big(r(s, a) +
\sum_{s' \in \Scal} P(s' \mid s, a) \max_{a' \in \Acal} \tilde{Q}^{k,
(+)}_{h+1}(s', a')\big) \geq 0,
\] we have
\[
\tilde{Q}^{k, (+)}_h(s, a) - \big(r(s, a) + \sum_{s' \in \Scal} P(s' \mid s, a) \max_{a' \in \Acal} \tilde{Q}^{k, (+)}_{h+1}(s', a')\big) \leq H - h + 1
\]
thus concluding the proof.
\end{proof}

Then, we show the optimism of $\tilde{V}^{k, (+)}(s) = \max_{a \in \Acal}
\tilde{Q}^{k, (+)}_h(s, a)$ for each $s, h, k \in \Scal \times [H] \times [K] \cup \{0\}$.
Define $V^*_h(s) = \max_{a \in \Acal} Q^*_h(s, a)$ for each $s \in \Scal$ and $h
\in [H]$ where 
$Q^*_h(s, a) = r(s, a) + \sum_{s' \in \Scal} P(s' \mid s, a) \max_{a' \in \Acal}
Q^*_{h+1}(s', a')$
 for each $s \in \Scal, a \in \Acal$ and $h \in [H]$ with the convention that $Q^*_{H+1}(s, a) = 0$ for each $s \in \Scal, a \in \Acal$. 
We show the following result.
\begin{lemma}
\label{lemma:optimism_V}
Under $\Ecal_2 \cap \Ecal_1$, for each $s \in \Scal, h \in [H], k \in [K] \cup
\{0 \}$, we have
\[
\tilde{V}^{k, (+)}_h(s) \geq V^*_h(s).
\]
\end{lemma}
\begin{proof}[Proof of \Cref{lemma:optimism_V}]
  The result for $k = 0$ comes from $V_h^* \leq H$. 
  For $k > 0$, assume $\Ecal_2 \cap \Ecal_1$ holds. We show the result by induction. The case $H+1$ is satisfied since $\tilde{V}^{k, (+)}_{H+1}(s) = 0$ and $V^*_{H+1}(s) = 0$ for each $s \in \Scal$. Then, assuming $\tilde{V}^{k, (+)}_{h+1}(s) \geq V^*_{h+1}(s)$ for each $s \in \Scal$, we have that for each $s \in \Scal$,
  \begin{align*}
    &\tilde{V}^{k, (+)}_h(s) - V^*_h(s) \\
    &= \max_{a \in \Acal} \tilde{Q}^{k, (+)}_h(s, a) - \max_{a \in \Acal} Q^*_h(s, a) \\
    &\geq \max_{a \in \Acal} r(s, a) + \sum_{s' \in \Scal} P(s' \mid s, a) \tilde{V}^{k, (+)}_{h+1}(s')
     \\&- \max_{a \in \Acal}\big( r(s, a) + \sum_{s' \in \Scal} P(s' \mid s, a) V^*_{h+1}(s')\big) && \text{(By $\Ecal_2$ and $\Ecal_1$ and Lemma~\ref{lemma:optimism})} \\ 
    &\geq \max_{a \in \Acal} r(s, a) + \sum_{s' \in \Scal} P(s' \mid s, a) V^*_{h+1}(s')
     \\&- \max_{a \in \Acal}\big( r(s, a) + \sum_{s' \in \Scal} P(s' \mid s, a) V^*_{h+1}(s')\big) && \text{(By $\Ecal_2$ and $\Ecal_1$ and induction hypothesis)} \\ 
     &\geq 0
  \end{align*}
\end{proof}

Finally, we show the regret bound. Assume $\Ecal_2 \cap \Ecal_1$ holds. 
We will first show by induction that for $k \in [K]$ and $h \in [H]$,
\begin{align}
&\EE[\tilde{V}^{k-1, (+)}_h(S^k_h) - V^{\pi^k}_h(S^k_h) \mid \Fcal_{k-1}, S^k_h] \notag \\
& \leq \EE[\sum_{i=h}^H b_{k-1}(S^k_i, \pi^k_i(S^k_i)) \mid \Fcal_{k-1}, S^k_h] \label{eq:value_bound} 
\end{align}

The above result is true for $h = H + 1$ since $\tilde{V}^{k-1, (+)}_{H+1}(s) =
0$ and  $V^{\pi^k}_{H+1}(s) = 0$ for each $s \in \Scal$.  
Then, assuming the
result holds for $h + 1$, we have that for each $k \in [K]$ that 
\begin{align*}
&\EE[\tilde{V}^{k-1, (+)}_h(S^k_h) - V^{\pi^{k}}_h(S^k_h) \mid \Fcal_{k-1}, S^k_h] \\
&= \EE[\max_{a \in \Acal} \tilde{Q}^{k-1, (+)}_h(S^k_h, a) - Q^{\pi^{k}}_h(S^k_h, \pi^{k}_h(S^k_h)) \mid \Fcal_{k-1}, S^k_h] \\
&= \EE[\tilde{Q}^{k-1, (+)}_h(S^k_h, \pi^{k}_h(S^k_h)) - Q^{\pi^{k}}_h(S^k_h, \pi^{k}_h(S^k_h)) \mid \Fcal_{k-1}, S^k_h] && \text{(By the definition of $\pi^{k}$)} \\
&\leq \EE[\tilde{V}^{k-1, (+)}_{h+1}(S^k_{h+1}) + b_{k-1}(S^k_h, \pi^{k}_h(S^k_h)) - V^{\pi^{k}}_{h+1}(S^k_{h+1}) \mid \Fcal_{k-1}, S^k_h] && \text{(By $\Ecal_2$ and $\Ecal_1$ and Lemma~\ref{lemma:optimism})} \\
&\leq \EE[ b_{k-1}(S^k_h, \pi^{k}_h(S^k_h)) + \EE[\sum_{i=h+1}^H b_{k-1}(S^k_i, \pi^k_i(S^k_i)) \mid \Fcal_{k-1}, S^k_{h+1}] \mid \Fcal_{k-1}, S^k_h] \\
&= \EE[\sum_{i=h}^H b_{k-1}(S^k_i, \pi^k_i(S^k_i)) \mid \Fcal_{k-1}, S^k_h]
\end{align*}

Then, the regret bound can be derived as follows. Given $s \in \Scal$, we have
\begin{align*}
&\sum_{k=1}^K \big(V^*_1(s) - V^{\pi^k}_1(s)\big) \\
&\leq \sum_{k=1}^K \big(\tilde{V}^{k-1, (+)}_1(s) - V^{\pi^k}_1(s)\big) && \text{(By Lemma~\ref{lemma:optimism_V})} \\
&\leq \sum_{k=1}^K \EE[\sum_{h=1}^H b_{k-1}(S^k_h, A^k_h) \ind{\Ecal_1 \cap \Ecal_2}\mid S^k_1=s] + 3 H^2 K \delta  && \text{(By \Cref{eq:value_bound})} \\
\end{align*}
We set $\delta = \frac1{3K}$ and then focus on $\sum_{k=1}^K \sum_{h=1}^H b_{k-1}(S^k_h, A^k_h)$.

\begin{align*}
&\sum_{k=1}^K \sum_{h=1}^H b_{k-1}(S^k_h, A^k_h) \\
&= \sum_{k=1}^K \sum_{h=1}^H \min\big(H - h + 1, \frac{18 S \beta_K(S^k_h, \pi^k_h(S^k_h)) H}{\max(1, N^{k-1}(S^k_h, \pi^k_h(S^k_h)) - 4 S \beta_K(S^k_h, \pi^k_h(S^k_h)))}\big)  \\
\end{align*}

Define 
\[
N^k_h(s, a) = N^{k-1}(s, a) + \sum_{i=1}^{h-1} \ind{S^k_i = s, A^k_i = a}
\]
for each $s \in \Scal, a \in \Acal, h \in [H]$ and $k \in [K]$. 
It holds that $N^{k-1}(s, a) \geq N^k_h(s, a) - H$ so we can write

\begin{align}
&\sum_{k=1}^K \sum_{h=1}^H b_{k-1}(S^k_h, A^k_h) \notag \\
&\leq \sum_{k=1}^K \sum_{h=1}^H \min\big(H - h + 1, \frac{18 S \beta_K(S^k_h, \pi^k_h(S^k_h)) H}{\max(1, N^k_h(S^k_h, \pi^k_h(S^k_h)) - 4 S \beta_K(S^k_h, \pi^k_h(S^k_h)) - H)}\big) \notag \\
&=\sum_{k=1}^K \sum_{h=1}^H \sum_{s \in \Scal} \sum_{a \in \Acal} \sum_{n \in [KH]} \ind{S^k_h=s, A^k_h=a, N^k_h(s, a)=n} \notag \\
&\min\big(H, \frac{18 S \beta_K(s, a) H}{\max(1, n - 4 S \beta_K(s, a) - H)}\big)  \label{eq:counting_bonuses}
\end{align}
Then, notice that given $s, a, n \in \Scal \times \Acal \times [KH]$,
$\ind{S^k_h=s, A^k_h=a, N^k_h(s, a)=n}$ is zero except for maybe a single couple $k, h$
we call $k(s, a, n), h(s, a, n)$, so we have
\begin{align*}
&\sum_{k=1}^K \sum_{h=1}^H b_{k-1}(S^k_h, A^k_h) \\
&\leq \sum_{s, a, n \in \Scal \times \Acal \times [KH]} \ind{S^{k(s, a, n)}_{h(s, a, n)}=s, A^{k(s, a, n)}_{h(s, a, n)}=a, N^{k(s, a, n)}_{h(s, a, n)}(s, a)=n} \\
&\min(H, \frac{18 S \beta_K(s, a) H}{\max(1, n - 4 S \beta_K(s, a) - H)}) \\
&\leq \sum_{s, a, n \in \Scal \times \Acal \times [KH]} \min(H, \frac{18 S \beta_K(s, a) H}{\max(1, n - 4 S \beta_K(s, a) - H)})  \\
&\leq 4 H S  \sum_{s\in \Scal, a \in \Acal} \beta_K(s, a) + H^2 SA  +  18 S \sum_{s, a \in \Scal \times \Acal} \beta_K(s, a) H \log(KH) \\
&\leq 22 HS \log(3KH) \sum_{s, a \in \Scal \times \Acal} \beta_K(s, a) + H^2 SA
\end{align*}
and then we can bound $\sum_{s, a \in \Scal \times \Acal} \beta_K(s, a)$ as follows.
\begin{align*}
  &\sum_{s \in \Scal, a \in \Acal}\beta_K(s, a)  \\
  &\leq \sum_{s \in \Scal, a \in \Acal} \sqrt{20 H \Upsilon_\epsilon^2 \sum_{h \in [H], i \in [K]} \ind{a = \pi^i_h(s)} \log(144K^3 S^2 A)} \\
  &+ 4 H SA \Upsilon_\epsilon \log(148K^3 S^2 A) \\
  &\leq S \sqrt{20 \Upsilon_\epsilon^2 AH^2K \log(148K^3 S^2 A)} + 4 H SA \Upsilon_\epsilon \log(148K^3 S^2 A) 
\end{align*}

The final regret bound is therefore given by
\begin{align*}
&\sum_{k \in [K]} \big(V^*_1(s) - V^{\pi^k}_1(s)\big) \\
&\leq 22 H^2S^2 \sqrt{20 \Upsilon_\epsilon^2 AK \log(148K^3 S^2 A)} \log(3KH) + 88 H^2S^2 A \Upsilon_\epsilon \log(148K^3 S^2 A) \log(3KH)\\
&+ H^2 SA + H^2 \\
&\leq 22 H^2S^2 \sqrt{20 \Upsilon_\epsilon^2 AK \log(148K^3 S^2 A)} \log(3KH)+ 92 H^2S^2 A \Upsilon_\epsilon \log(148K^3 S^2 A) \log(3KH)
\end{align*}

\end{proof}
\newpage
\section*{NeurIPS Paper Checklist}

\begin{enumerate}

\item {\bf Claims}
    \item[] Question: Do the main claims made in the abstract and introduction accurately reflect the paper's contributions and scope?
    \item[] Answer: \answerYes{}
    \item[] Justification: The abstract and introduction clearly state the minimax regret bounds claimed in the paper. All claims are supported by formal theorems (\Cref{thm:lower_bound}, \Cref{thm:upper_bound}, \Cref{cor:lower_bound}, \Cref{thm:upper_bound_rl}) with complete proofs, and the scope (time-homogeneous episodic tabular RL under non-interactive $\epsilon$-LDP) is precisely defined.

\item {\bf Limitations}
    \item[] Question: Does the paper discuss the limitations of the work performed by the authors?
    \item[] Answer: \answerYes{}
    \item[] Justification: The conclusion section discusses the remaining gap of $HS$ between lower and upper bounds for the high-privacy regime ($\epsilon \leq 1$), the sub-optimality of the upper bound in the low-privacy regime, and the conjecture that the $S^2$ dependency is unimprovable for model-based algorithms.

\item {\bf Theory assumptions and proofs}
    \item[] Question: For each theoretical result, does the paper provide the full set of assumptions and a complete (and correct) proof?
    \item[] Answer: \answerYes{}
    \item[] Justification: All theorems and corollaries are numbered, their assumptions are explicitly stated, and complete proofs are provided in the appendix (\Cref{app:lower_bound,app:upper_bound,app:upper_bound_rl}). Proof sketches in the main body provide intuition for the full appendix proofs.

    \item {\bf Experimental result reproducibility}
    \item[] Question: Does the paper fully disclose all the information needed to reproduce the main experimental results of the paper to the extent that it affects the main claims and/or conclusions of the paper (regardless of whether the code and data are provided or not)?
    \item[] Answer: \answerNA{}
    \item[] Justification: The paper is purely theoretical and contains no experiments.

\item {\bf Open access to data and code}
    \item[] Question: Does the paper provide open access to the data and code, with sufficient instructions to faithfully reproduce the main experimental results, as described in supplemental material?
    \item[] Answer: \answerNA{}
    \item[] Justification: The paper is purely theoretical and contains no experiments requiring code or data.

\item {\bf Experimental setting/details}
    \item[] Question: Does the paper specify all the training and test details (e.g., data splits, hyperparameters, how they were chosen, type of optimizer) necessary to understand the results?
    \item[] Answer: \answerNA{}
    \item[] Justification: The paper is purely theoretical and contains no experiments.

\item {\bf Experiment statistical significance}
    \item[] Question: Does the paper report error bars suitably and correctly defined or other appropriate information about the statistical significance of the experiments?
    \item[] Answer: \answerNA{}
    \item[] Justification: The paper is purely theoretical and contains no experiments.

\item {\bf Experiments compute resources}
    \item[] Question: For each experiment, does the paper provide sufficient information on the computer resources (type of compute workers, memory, time of execution) needed to reproduce the experiments?
    \item[] Answer: \answerNA{}
    \item[] Justification: The paper is purely theoretical and contains no experiments.
    
\item {\bf Code of ethics}
    \item[] Question: Does the research conducted in the paper conform, in every respect, with the NeurIPS Code of Ethics \url{https://neurips.cc/public/EthicsGuidelines}?
    \item[] Answer: \answerYes{}
    \item[] Justification: 

\item {\bf Broader impacts}
    \item[] Question: Does the paper discuss both potential positive societal impacts and negative societal impacts of the work performed?
    \item[] Answer: \answerNA{}
    \item[] Justification: This is foundational theoretical research on privacy-preserving reinforcement learning. The work contributes to understanding fundamental limits of private learning, with no direct path to negative applications.
    
\item {\bf Safeguards}
    \item[] Question: Does the paper describe safeguards that have been put in place for responsible release of data or models that have a high risk for misuse (e.g., pre-trained language models, image generators, or scraped datasets)?
    \item[] Answer: \answerNA{}
    \item[] Justification: The paper releases no models, datasets, or other assets that pose a risk of misuse.

\item {\bf Licenses for existing assets}
    \item[] Question: Are the creators or original owners of assets (e.g., code, data, models), used in the paper, properly credited and are the license and terms of use explicitly mentioned and properly respected?
    \item[] Answer: \answerNA{}
    \item[] Justification: The paper does not use any external code, datasets, or models. All prior works are properly cited.

\item {\bf New assets}
    \item[] Question: Are new assets introduced in the paper well documented and is the documentation provided alongside the assets?
    \item[] Answer: \answerNA{}
    \item[] Justification: The paper introduces no new datasets, models, or code assets.

\item {\bf Crowdsourcing and research with human subjects}
    \item[] Question: For crowdsourcing experiments and research with human subjects, does the paper include the full text of instructions given to participants and screenshots, if applicable, as well as details about compensation (if any)?
    \item[] Answer: \answerNA{}
    \item[] Justification: The paper does not involve crowdsourcing or research with human subjects.

\item {\bf Institutional review board (IRB) approvals or equivalent for research with human subjects}
    \item[] Question: Does the paper describe potential risks incurred by study participants, whether such risks were disclosed to the subjects, and whether Institutional Review Board (IRB) approvals (or an equivalent approval/review based on the requirements of your country or institution) were obtained?
    \item[] Answer: \answerNA{}
    \item[] Justification: The paper does not involve crowdsourcing or research with human subjects.

\item {\bf Declaration of LLM usage}
    \item[] Question: Does the paper describe the usage of LLMs if it is an important, original, or non-standard component of the core methods in this research? Note that if the LLM is used only for writing, editing, or formatting purposes and does \emph{not} impact the core methodology, scientific rigor, or originality of the research, declaration is not required.
    \item[] Answer: \answerNA{}
    \item[] Justification: ChatGPT 5.5 was used to brainstorm proof strategies, verifying proposed algorithms and proofs, proofreading and polishing the manuscript. 
    The authors verified the correctness and originality of all content.

\end{enumerate}

\end{document}